%% file: main.tex
\PassOptionsToPackage{table}{xcolor}
\documentclass[11pt]{tea}
\microtypesetup{expansion=false}

\usepackage[utf8]{inputenc}
\usepackage{amsmath}
\usepackage{nicefrac}
\usepackage{enumitem}
\usepackage{algorithm}
\usepackage{algpseudocode}
\usepackage{fvextra}
\usepackage{makecell}
\usepackage{tabularx}
\usepackage{soul}
\usepackage{wrapfig}
\usepackage{needspace}

\newcommand{\ours}{\texorpdfstring{\textnormal{\scshape VisionSI}}{VisionSI}}

\title{\ours: Do Self-Improving Vision-Language Agents Really Learn to See? Diagnosing and Mitigating Evidence Drift}

\author[1,2]{Ruinan Jin}
\author[1,2]{Gexin Huang}
\author[1]{Jiaqi Liu}
\author[3]{Qiong Zhang}
\author[1,2]{Xiaoxiao Li}

\affiliation[1]{The University of British Columbia}
\affiliation[2]{Vector Institute}
\affiliation[3]{\mbox{Mohamed bin Zayed University of Artificial Intelligence (MBZUAI)}}

\hypersetup{pdfauthor={Ruinan Jin, Gexin Huang, Jiaqi Liu, Qiong Zhang, Xiaoxiao Li}}
\hypersetup{pdftitle={VisionSI: Do Self-Improving Vision-Language Agents Really Learn to See? Diagnosing and Mitigating Evidence Drift}}
\abstract{\input{sections/abstract}}
\metadata[Contact]{\email{ruinanjin@alumni.ubc.ca}}
\metadata[Project page]{\url{https://github.com/Nanboy-Ronan/VISIONSI}}

\begin{document}
\maketitle

\input{sections/introduction}
\input{sections/related_work}
\input{sections/method}
\input{sections/experiments}
\input{sections/conclusion}

\input{sections/statements}
\bibliographystyle{plainnat}
\bibliography{references}

\clearpage
\beginappendix
\input{sections/appendix}
\end{document}

%% file: sections/abstract.tex
Self-improvement (SI) and recursive self-improvement (RSI) enable vision-language agents to evolve their harnesses, including prompts, workflows, and reusable skills, without updating model weights. But do these agents learn to see better, or merely to answer better? We identify \textbf{evidence drift}: answer-only harness optimization can improve answer accuracy while degrading the localization of reported visual evidence. Across three vision-language models (VLMs), 3.8 percentage points of GEPA's 7.3-point accuracy gain persist under whole-image blurring, suggesting that answer-only optimization can reward textual shortcuts rather than improved visual observation. To mitigate evidence drift, we introduce \ours{}, which integrates reliability-aware visual credit assignment into existing SI and RSI methods without bounding-box supervision. Whole-image and region-specific visual interventions estimate the reliability of model-generated visual references, determining how much credit is assigned to reference agreement during harness selection. An auxiliary Visual-Coherent Batch Sampler facilitates the learning of reusable visual instructions. Across four VLMs, five visual domains, and ten SI and RSI baselines, integrating \ours{} into GEPA and Hyperagents improves average answer accuracy by 8.3 and 12.3 percentage points, respectively, with corresponding gains of 10.8 and 8.3 points in mean evidence IoU. These findings highlight the importance of visual grounding beyond answer accuracy in self-improving multimodal agents.

%% file: sections/introduction.tex
\section{Introduction}
\label{sec:intro}

Vision-language models can produce plausible text that conflicts with the image when learned textual shortcuts override visual evidence~\citep{leng2024mitigating}. Many hallucination methods therefore seek to restore the influence of visual evidence on a single prediction. Visual Contrastive Decoding, for example, contrasts predictions from original and perturbed images to suppress outputs that survive without reliable visual support~\citep{leng2024mitigating}. The same failure becomes more consequential when a VLM is embedded in a self-improving agent. A hallucinated answer affects one execution. A misguided visual instruction can be retained and reused across many later executions.

\Needspace{.45\textheight}
\begin{wrapfigure}{r}{.48\linewidth}
  \centering
  \includegraphics[width=\linewidth]{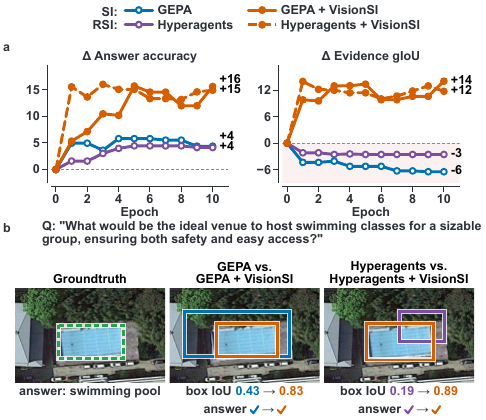}
  \caption{\textbf{Answer improvement can conceal evidence drift.} (a) GEPA (SI) and Hyperagents (RSI) improve answer accuracy while the IoU of their reported evidence boxes decreases. Adding \ours{} improves both accuracy and localization in this setting. (b) An EarthReason test example with Qwen3.5-2B. All four harnesses return the correct answer, but \ours{} localizes the swimming pool more accurately in both SI and RSI.}
  \label{fig:teaser}
\end{wrapfigure}
Self-improving agents turn execution feedback into persistent changes around a frozen model. A task agent, which we call the \emph{executor}, performs tasks using a frozen model and a \emph {harness} comprising prompts, workflows, reusable skills, etc. A separate procedure, called the \emph{improver}, revises this harness through prompt optimization, context construction, or harness search~\citep{agrawal2026gepa,yuksekgonul2024textgrad,zhang2026agentic,ouyang2026reasoningbank,yang2026skillopt,lee2026meta}. In \emph{self-improvement} (SI), the improver remains fixed while the executor's harness changes. In \emph{recursive self-improvement} (RSI), earlier revisions can also influence how later revisions are produced, sometimes by modifying the improver itself~\citep{xia2026rrsi,zhang2026darwin,zhang2026hyperagents}. Both settings usually select revisions by answer correctness. However, a correct answer may result from better visual observation, a learned textual shortcut, or a change in response style. Answer feedback distinguishes none of them. It can teach an agent what to answer while remaining silent about which evidence the agent reports. This raises a question: \ul{\emph{Is answer correctness a sufficient optimization signal for self-improving vision-language agents?}}

To examine whether optimizing answer correctness also improves reported evidence localization, we study five domains with different visual evidence requirements: autonomous driving~\citep{corbiere2025retrieval}, robot manipulation~\citep{yang2026vista}, remote sensing~\citep{li2025segearth}, controlled compositional reasoning~\citep{li2023super}, and mobile interface understanding~\citep{hsiao2025screenqa}. The relevant evidence ranges from small road signs and manipulation targets to large geographic regions, synthetic objects, and interface elements. For each question, the agent reports both an answer and bounding boxes identifying the image regions it considers relevant, allowing us to evaluate answer correctness and reported evidence localization separately. This diversity allows us to test whether reliability-aware visual credit can guide harness improvement across tasks requiring different types and scales of visual evidence.

We first analyze GEPA because it explicitly selects candidate harness revisions by answer correctness. As shown in Figure~\ref{fig:teaser}, gains in answer accuracy are only weakly coupled with changes in evidence localization. We define \emph{evidence drift} as a retained harness revision that improves answer accuracy while degrading the localization of its reported boxes. Figure~\ref{fig:teaser} shows this failure. Evidence drift exposes a missing credit signal: answer feedback assigns equal credit to two correct revisions even when only one improves reported localization.

We introduce \ours{} to mitigate evidence drift by integrating reliability-aware visual credit assignment into existing SI and RSI methods without box supervision. The visual credit assigned to agreement with a model-generated reference depends on two quantities: (i) how much the model's support for the correct answer depends on the image, and (ii) how much of that dependence is captured by the reference. Blurring the entire image measures image reliance, while blurring only the reference measures how much of that reliance it captures. We combine these measurements into a reliability weight that controls how much reference overlap contributes when ranking harness revisions.
An auxiliary Visual-Coherent Batch Sampler uses frozen image-question embeddings to group questions with compatible visual needs and restrict where each learned visual instruction is reused. The VLM remains frozen, annotated boxes are never used during optimization, and inference requires no visual intervention. We apply this credit signal to GEPA for self-improvement (SI) and Hyperagents for recursive self-improvement (RSI), where earlier revisions can also change how later revisions are produced~\citep{xia2026rrsi,zhang2026darwin,zhang2026hyperagents}. Both hosts select among candidate harness revisions, allowing us to directly evaluate visual credit without changing their proposal mechanisms.


In the five-domain matched evaluation, \ours{} improves average answer accuracy and gIoU over both GEPA and the RSI host Hyperagents. Further analysis shows that visual credit drives the localization gain and that additional host compute alone does not reproduce it.

Our contributions are threefold:
\begin{itemize}[leftmargin=*,nosep]
    \item We identify evidence drift: optimizing answer correctness can improve answers while degrading reported evidence localization, revealing a limitation of answer correctness as an optimization signal.
    \item We propose \ours{}, a credit-assignment method that supplements answer correctness with reliability-aware visual credit for harness selection, without bounding-box supervision.
    \item We evaluate ten self-improvement methods with four VLMs across five visual domains. Adding \ours{} to GEPA and Hyperagents improves average answer accuracy and evidence localization over the corresponding hosts.
\end{itemize}

%% file: sections/related_work.tex
\section{Related Work}
\label{sec:related}
\paragraph{Self-Improving and Recursive Self-Improving Agents.}
A model harness is the prompts, workflow, memory, and skills around a frozen model. Harness self-improvement changes these components rather than the model weights. In self-improvement (SI), a fixed improver runs the current harness, uses feedback from its executions, and writes a revised harness for future tasks. Existing methods retain verbal feedback or revise instructions, skills, memories, contexts, and workflows~\citep{shinn2023reflexion,
pryzant2023automatic,yuksekgonul2024textgrad,agrawal2026gepa,yang2026skillopt,zhang2026agentic,
ouyang2026reasoningbank,hu2025automated,zhang2025aflow}.

Recursive self-improvement (RSI) also lets an accepted harness and its execution history influence how later revisions are produced. RRSI carries this state across generations, while DGM and Hyperagents can additionally change parts of the improver itself~\citep{xia2026rrsi,zhang2026darwin,zhang2026hyperagents}. Thus, SI changes the task agent
with a fixed improver, whereas RSI allows earlier changes to shape how later agents are built. Both families were
developed primarily for language and code tasks, where feedback checks an answer or program and retained experience is text or code. Applied to a vision-language agent, this feedback can select a better answer without showing whether the agent used the right image content. We therefore evaluate both answers and reported image regions under SI and RSI.

\paragraph{Hallucination and Textual Shortcut Learning in Vision-Language Models and Agents.}
Prior work shows that vision-language systems can learn
textual shortcuts instead of using visual evidence. Studies of VQA
demonstrate that question-answer correlations can produce correct predictions
without adequate image understanding~\citep{goyal2017making,agrawal2018don}, reflecting the broader phenomenon of shortcut
learning~\citep{geirhos2020shortcut}.
For large vision-language models, POPE and
HallusionBench evaluate object hallucinations and
failures involving language and visual evidence~\citep{li2023evaluating,guan2024hallusionbench}. Methods such as Visual Contrastive Decoding and multimodal
mutual-information decoding mitigate hallucinations by
strengthening the influence of visual information during
generation~\citep{leng2024mitigating,favero2024multi}.
Recent work also investigates whether VLMs can iteratively correct
their own semantic grounding errors~\citep{liao2025can}. These studies primarily examine individual predictions, decoding strategies, or inference-time corrections. In contrast, answer feedback can encode a textual shortcut into the persistent instructions and skills of a self-improving agent, allowing it to influence later predictions. We study whether such persistent adaptation improves reported visual evidence alongside answer accuracy. Our method uses image interventions to assign reliability-aware visual credit to harness revisions without annotated bounding-box supervision.

%% file: sections/method.tex
\section{Method}
\label{sec:method}
We first define the problem settings in Sec.~\ref{sec:prelim}, investigate why answer feedback can reward updates that lead to evidence drift in Sec.~\ref{sec:analysis}, and introduce reliability-aware visual credit to mitigate the evidence drift in Sec.~\ref{sec:framework}.

\subsection{Problem Setting}
\label{sec:prelim}
We consider a visual agent built from a frozen VLM and a harness $H$. The harness contains the instructions, workflow, and skills that determine how the model is used. Given an image and question $x_i=(I_i,q_i)$, the frozen VLM executes the harness to produce a response. We call this frozen task agent the \emph{executor}. Self-improvement evolves the executor's harness while keeping the VLM fixed.

The harness is improved on a training set of image-question pairs with correct answers $y_i$. The \emph{improver} proposes candidate harnesses, runs them on the training questions, and selects which revisions to retain. In SI, the improver remains fixed while the harness changes. In RSI, a retained harness and its execution history influence later revisions. Some RSI methods also revise the improver itself. In both settings, training experience is converted into a harness that can be reused on future questions. We call the underlying SI or RSI algorithm the \emph{host} method. \ours{} integrates into its credit-assignment process to provide reliability-aware visual feedback.

Existing SI and RSI methods were designed for language and code tasks, where a revision is useful when it produces a better answer or completes a task. On a visual task, answer feedback alone cannot reveal whether the reported evidence stays on the relevant regions. We make this distinction measurable through a shared output interface. For each question, the executor returns an answer $a_H(x_i)$ and boxes $B_H(x_i)$ that report the relevant image regions. The fixed interface supports a one-call workflow that predicts both fields from the full image and a two-stage workflow that also supplies region crops when producing the answer. In this workflow, the reported boxes also select the crops the executor sees when it answers. Training still uses only the correct answer $y_i$. On held-out testing VQAs, we evaluate answer correctness $A_i(H)\in\{0,1\}$ and the mean per-question IoU of the reported boxes. A revision can therefore be evaluated by both what the agent answers and which image regions it reports. The next section shows that the two can move in opposite directions, a failure we call \emph{evidence drift}.

\subsection{Diagnosing Evidence Drift under Answer-Only Optimization}
\label{sec:analysis}

We next examine why answer-only credit assignment may lead to evidence drift. Answer-only credit may improve accuracy through better visual observation or through image-independent cues, including textual shortcuts retained in the harness. Such cues are not necessarily harmful, but answer feedback alone cannot determine whether an improvement is accompanied by better evidence localization. Consequently, a harness may gain accuracy while its reported boxes drift away from ground-truth evidence regions. We examine this failure by comparing the initial harness with the harness returned by GEPA under progressively stronger whole-image blur. If an answer gain persists under severe visual degradation, it cannot be attributed solely to improved image observation. In this paper, ground-truth bounding boxes are used solely for evaluating the localization accuracy of the model-reported evidence using IoU.

\Needspace{.56\textheight}
\begin{wrapfigure}{r}{.47\linewidth}
  \centering
  \includegraphics[width=\linewidth]{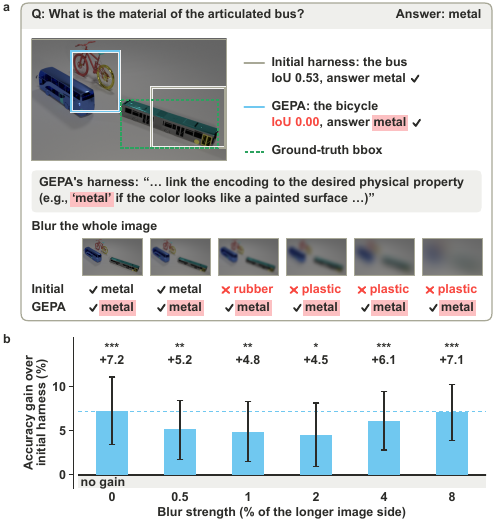}
\caption{\textbf{GEPA retains its answer gain under image blurring.}
(a) On the original image, both harnesses correctly answer ``metal'', but GEPA localizes the bicycle instead of the articulated bus. This is because the answer token ``metal'' directly appears in GEPA's optimized harness.
(b) GEPA consistently outperforms the initial harness in answer accuracy across all blur strengths, indicating that its accuracy gains persist even as visual evidence is blurred. Asterisks indicate the statistical significance of accuracy differences between GEPA and the initial harness at each blur strength, assessed using exact McNemar tests (${}^{*}p<0.05$, ${}^{**}p<0.01$, ${}^{***}p<0.001$).}
\label{fig:why_drift}
\end{wrapfigure}
\paragraph{Answer gains persist under heavy blur.}
We blur the whole image to examine how much of GEPA's answer gains depend on visual signals. For Qwen3.5-2B, GEPA's accuracy gain over the initial harness is 7.2 points on the original images and remains at 7.1 points under the strongest blur. Its advantage remains significant at every blur strength, as shown in Figure~\ref{fig:why_drift}b. Across the three VLMs in this controlled analysis, 3.8 points of GEPA's 7.3-point gain remain after whole-image blurring.

These results show that part of the improvement selected by answer correctness persists even when visual information is heavily degraded. Together with the localization errors in Figure~\ref{fig:why_drift}a, they expose a limitation of answer-only credit: a higher answer score does not establish that a harness revision makes better use of the image.

\paragraph{Textual shortcuts decouple answers from visual evidence.}
The example in Figure~\ref{fig:why_drift}a shows how this happens. It asks for the material of an articulated bus,
and both harnesses answer metal on the image. The initial harness (zero-shot the original agent) boxes the bus with IoU 0.53, while GEPA boxes the
bicycle next to it with IoU 0.00. Once the image is blurred beyond 0.5\%, the initial harness answers rubber or plastic,
and GEPA keeps answering metal at every strength. This is because GEPA's optimized harness gives ``metal'' as an example answer for material questions. Namely, it memorized the answer rather than learning the observation patterns. Answer reward assigns equal credit to this shortcut and to a revision that uses the image. It can therefore optimize a harness whose answers improve while its reported evidence drifts away from the regions
relevant to the question. \emph{Answer accuracy is therefore an incomplete credit signal for a visual improver.} It rewards a correct answer without checking whether its reported evidence stays on the correct regions. 

\subsection{\ours{}: Reliability-Aware Visual Credit}
\label{sec:framework}

\begin{figure}[tbp]
  \centering
  \includegraphics[width=.90\textwidth]{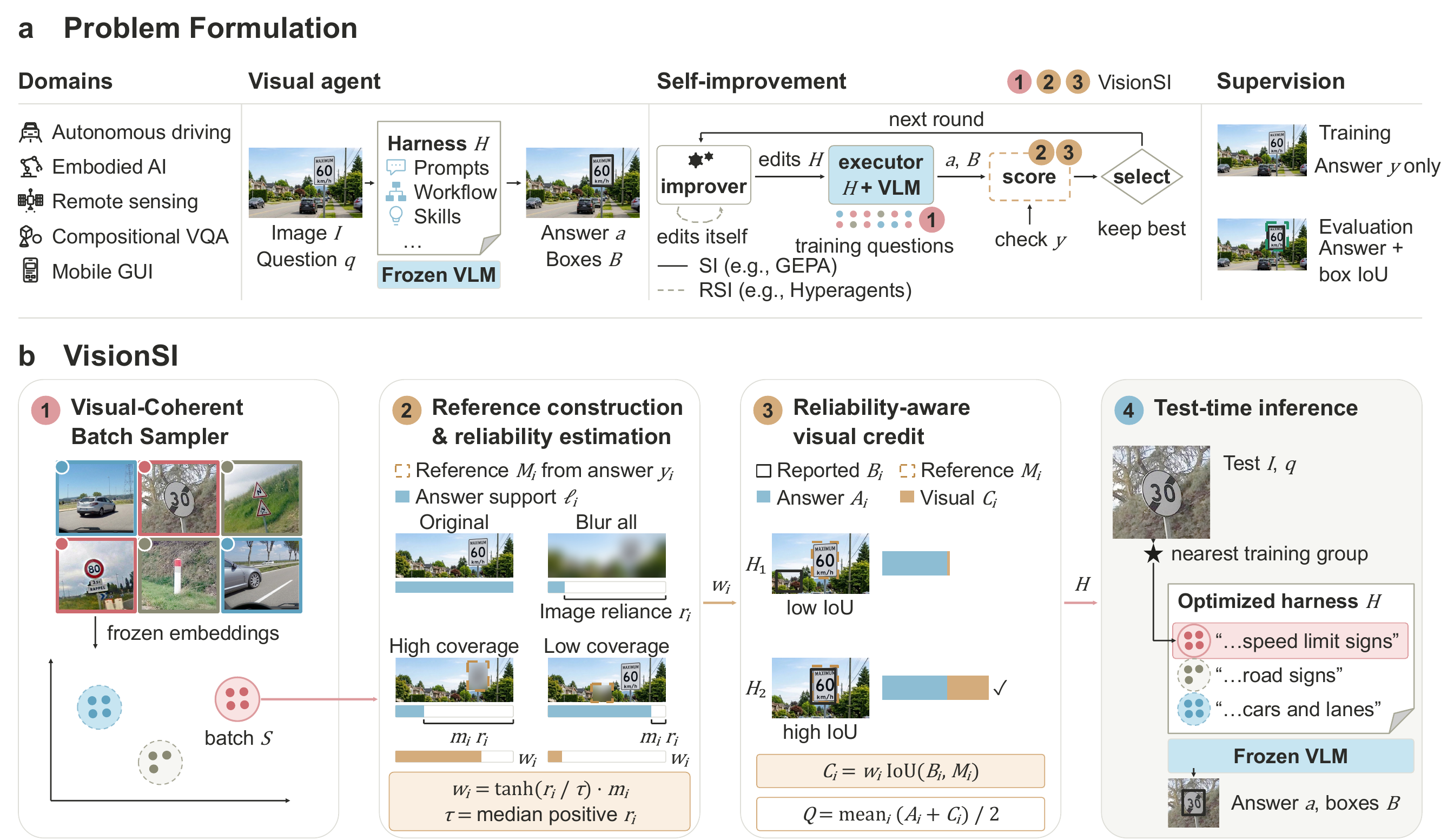}
\caption{\textbf{Overview of \ours{}.}
(a) A frozen VLM executes a harness to produce an answer and evidence boxes. An SI or RSI host evolves the harness using answer labels, while annotated boxes are reserved for analysis and evaluation. (b) \ours{} uses visual interventions to estimate image reliance and how much of that reliance a model-generated reference captures. These estimates determine a reliability weight that scales the visual credit assigned to reference agreement during harness improvement. The Visual-Coherent Batch Sampler groups image-question pairs with similar visual patterns to strengthen the learning signal for reusable visual instructions.}
\label{fig:overview}
\end{figure}
\Needspace{.33\textheight}
\begin{wrapfigure}{r}{.44\linewidth}
  \captionsetup{font=footnotesize,skip=4pt}
  \vspace{-0.6\baselineskip}
  \centering
  \includegraphics[width=.93\linewidth]{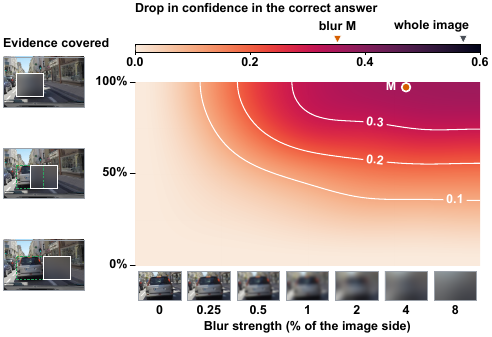}
\caption{\textbf{Visual interventions help evaluate predicted evidence boxes.}
We shift the blur window away from the annotated region (rows) and increase blur strength (columns). At 4\% blur, masking the predicted evidence box $M$ (orange) reduces confidence in the correct answer by nearly as much as masking all annotated evidence.}
\label{fig:motivation}
\end{wrapfigure}
Evidence drift reveals a limitation of answer-only optimization: improved answer accuracy does not guarantee better localization of the reported visual evidence. Answer correctness alone therefore cannot determine whether a revision better aligns its answer with its reported visual evidence. \ours{} addresses this ambiguity by augmenting the improver's credit-assignment process. We reuse the whole-image intervention from Sec.~\ref{sec:analysis} not as an objective, but to decide when a model-generated reference is reliable enough to receive credit. A candidate box receives credit only when it overlaps a region that carries image-dependent support for the correct answer. Figure~\ref{fig:overview}b shows the procedure.

Given the correct answer $y_i$, the frozen VLM predicts an answer-conditioned reference region $M_i$ for training
example $i$. Since $M_i$ is model-generated, overlap with it should not receive credit unconditionally. We first
determine whether the answer depends on the image and whether $M_i$ captures that dependence.
Let $\ell_i(V)$ denote the frozen VLM's log-odds for $y_i$ under image view $V$. We compare the original image $I_i$
with a fully blurred image $\bar I_i$ and with $I_i^{\setminus M_i}$, in which only $M_i$ is blurred:
\begin{equation}
\begin{aligned}
 r_i&=\left[\ell_i(I_i)-\ell_i(\bar I_i)\right]_+,\\
 m_i&=\operatorname{clip}_{[0,1]}\!\left(
 \frac{\ell_i(I_i)-\ell_i(I_i^{\setminus M_i})}{r_i}
 \right) \quad \text{for } r_i>0.
\end{aligned}
 \label{eq:deletion_tests}
\end{equation}
The image reliance $r_i$ applies the whole-image intervention from Figure~\ref{fig:why_drift}, which limits credit to answers that depend on the image. The reference coverage $m_i$ measures the fraction of this reliance removed by blurring $M_i$. Figure~\ref{fig:motivation} shows that this localized intervention tracks the removal of annotated
evidence. We set $m_i=0$ when $r_i=0$ or $M_i$ is empty.

We combine image reliance and reference coverage into the reliability weight
\begin{equation}
 w_i=\tanh\!\left(\frac{r_i}{\tau}\right)m_i,
 \label{eq:credit_weight}
\end{equation}
where $\tau$ is the median positive $r_i$ on the training set. The first factor suppresses visual credit when the
answer has little dependence on the image. The second suppresses credit when $M_i$ misses the visual support that
remains. Agreement with $M_i$ therefore matters only when both conditions hold.

For candidate harness $H$, let $A_i(H)$ denote answer correctness and let $B_H(x_i)$ denote its boxes. We assign the
candidate visual credit
\begin{equation}
 C_i(H)=w_i\operatorname{IoU}\!\left(B_H(x_i),M_i\right).
 \label{eq:visual_credit}
\end{equation}
For a training batch $\mathcal S$, the improver scores the candidate by
\begin{equation}
 Q(H,\mathcal S)=\frac{1}{|\mathcal S|}\sum_{i\in\mathcal S}
 \frac{A_i(H)+C_i(H)}{2}.
 \label{eq:host_credit}
\end{equation}
When $w_i=0$, the ranking is identical to answer-only credit. As $w_i$ increases, candidates receive more credit for boxes that agree with a reliable reference. We integrate this credit function into GEPA~\citep{agrawal2026gepa} and Hyperagents~\citep{zhang2026hyperagents}, augmenting their credit-assignment process with reliability-aware visual feedback. Their proposal mechanisms and optimization budgets remain unchanged.

Visual credit is computed per example, but a reusable harness update must be learned from a batch. Mixing questions that require unrelated visual observations weakens their shared learning signal, as shown by the batch-composition analysis in the Appendix. We therefore use an auxiliary Visual-Coherent Batch Sampler. It groups training questions with frozen image-question embeddings and draws each batch from one group. Within each group, the improver proposes one short visual instruction and retains it only if visual credit improves without reducing answer correctness. This local acceptance rule applies to the group-specific instruction, whereas Eq.~\ref{eq:host_credit} scores complete harness revisions proposed by the host. The same groups determine where each retained instruction is reused. The sampler introduces no additional supervision. Implementation details are provided in the Appendix.

At inference, the input receives the visual instruction of its nearest group, and the optimized harness runs once.

%% file: sections/experiments.tex
\section{Experiments}
\label{sec:experiments}

We first describe the experimental settings in Sec.~\ref{sec:setup}, then evaluate \ours{} with GEPA for SI and Hyperagents for RSI in Sec.~\ref{sec:results}. Finally, we analyze the changes made during self-improvement and the contributions of our method's components in Sec.~\ref{sec:discussion}.

\input{sections/baseline_table}

\subsection{Experimental Settings}
\label{sec:setup}
\paragraph{Datasets.}
We evaluate five visual domains where both answer correctness and visual evidence matter: robot manipulation in UMI-VQA~\citep{yang2026vista}, compositional reasoning in Super-CLEVR~\citep{li2023super}, remote sensing in EarthReason~\citep{li2025segearth}, autonomous driving in DrivingVQA~\citep{corbiere2025retrieval}, and mobile interface understanding in ScreenQA~\citep{hsiao2025screenqa}. These domains require harnesses to prioritize different visual cues, from object attributes and spatial relations to geospatial structures, traffic scenes, and interface elements. This diversity allows us to test whether reliability-aware visual credit can guide harness optimization across distinct visual reasoning tasks. Representative images are provided in the supplementary material.


\paragraph{Models and baselines.}
We evaluate four compact and open-weight VLMs: Qwen3VL-8B~\citep{bai2025qwen3}, Gemma4-E4B~\citep{team2026gemma}, Qwen3.5-2B~\citep{team2026qwen3}, and InternVL3.5-8B~\citep{wang2025internvl3}. These models span different families and capacities, allowing us to examine whether visual credit improves harness optimization across model backbones. Their moderate inference costs also make repeated SI and RSI experiments computationally feasible. We compare with six SI baselines and four RSI baselines with details in Appendix. All methods use the same frozen VLM, training split, and agent interface. All methods are supervised only by answer labels, without ground-truth bounding boxes. As mentioned in Sec.~\ref{sec:intro}, we instantiate \ours{} in GEPA~\citep{agrawal2026gepa} for SI and Hyperagents~\citep{zhang2026hyperagents} for RSI to evaluate visual credit under both fixed and evolving improvement mechanisms. In both hosts, \ours{} augments the credit-assignment process without changing how candidate harness evolvements are proposed.

\paragraph{Metrics.}
We report answer accuracy and two localization metrics. gIoU is the mean per-question IoU between the unions of
predicted and annotated boxes. cIoU is the summed intersection divided by the summed
union across questions. 

\subsection{Experimental Results}
\label{sec:results}
\paragraph{\ours{} improves average answer accuracy and evidence localization.}
Table~\ref{tab:main} reports the main comparison across all settings. \emph{In SI}, GEPA averages 58.7 accuracy, 36.4 gIoU, and 31.2 cIoU. Adding \ours{} increases these scores to 67.0, 47.2, and 41.4, yielding gains of 8.3, 10.8, and 10.2 points, respectively. \emph{In RSI}, Hyperagents averages 56.0 accuracy, 38.8 gIoU, and 32.6 cIoU, while \ours{} achieves 68.3, 47.1, and 41.1, corresponding to gains of 12.3, 8.3, and 8.4 points. Beyond these average improvements, \ours{} ranks among the top two methods on most reported metrics within its respective SI or RSI group. These gains span diverse visual domains.

\subsection{Further Discussion}
\label{sec:discussion}

\paragraph{Answer gains and evidence localization.}
On UMI-VQA with Qwen3.5-2B, GEPA raises test accuracy from 17.1\% to 29.0\%, while mean box IoU (gIoU) changes only from 10.5 to 10.9. Hyperagents raises accuracy to 17.6\%, while gIoU falls to 10.1. Thus, both hosts improve answer accuracy, but neither substantially improves the localization of its reported boxes. Adding \ours{} improves both metrics for both hosts. Across all four models and five domains, \ours{} improves gIoU in 38 of the 40 matched host comparisons and improves both accuracy and gIoU in 31.

\begin{table}[htbp]
\caption{\textbf{Ablation of visual credit and batch sampler.} Results are on the UMI-VQA test set with Qwen3.5-2B, in percent.}
\label{tab:ablation}
\centering
\scriptsize
\setlength{\tabcolsep}{2.6pt}
\renewcommand{\arraystretch}{0.94}
\setlength{\aboverulesep}{1pt}\setlength{\belowrulesep}{1pt}
\begin{tabular*}{\linewidth}{@{\extracolsep{\fill}}lcccccc@{}}
\toprule
 & \multicolumn{3}{c}{GEPA} & \multicolumn{3}{c}{Hyperagents} \\
\cmidrule(lr){2-4} \cmidrule(lr){5-7}
Setting & Acc. & gIoU & cIoU & Acc. & gIoU & cIoU \\
\midrule
\textit{Initial harness} & 17.1 & 10.5 & 13.1 & 17.1 & 10.5 & 13.1 \\
\ours{} & 34.8 & 37.3 & 39.5 & 38.6 & 37.0 & 38.4 \\
\midrule
w/o visual credit & 32.4 & 16.6 & 16.2 & 34.3 & 12.4 & 15.1 \\
w/o batch sampler & 31.9 & 31.7 & 26.3 & 31.4 & 35.9 & 34.9 \\
w/o both & 29.0 & 10.9 & 14.2 & 17.6 & 10.1 & 12.0 \\
\bottomrule
\end{tabular*}
\end{table}

\paragraph{Component analysis.}
Table~\ref{tab:ablation} shows how visual credit and visual coherent batch sampler affect performance on UMI-VQA with Qwen3.5-2B. Removing visual credit reduces gIoU from 37.3 to 16.6 for GEPA and from 37.0 to 12.4 for Hyperagents, while accuracy falls by 2.4 and 4.3 points, respectively. Removing visual coherent batch sampler reduces accuracy by 2.9 points for GEPA and 7.2 points for Hyperagents, and reduces cIoU by 13.2 and 3.5 points, respectively. For both hosts, removing visual credit causes the larger drop in gIoU, while removing visual-coherent batch sampler causes the larger drop in accuracy.

\Needspace{.35\textheight}
\begin{wrapfigure}{r}{.44\linewidth}
  \centering
  \includegraphics[width=\linewidth]{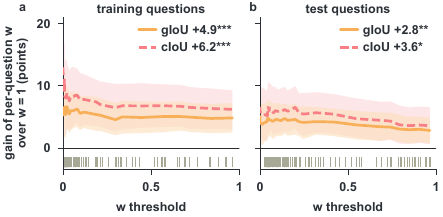}
  \caption{\textbf{Effect of reliability weighting.} Each point shows the gain from per-question weights $w_i$ over uniform weights $w_i=1$, evaluated on (a) training and (b) test questions with $w_i\le t$. The rightmost point includes all questions. Bottom ticks mark individual $w_i$ values. Results are on UMI-VQA with Qwen3.5-2B, averaged over GEPA and Hyperagents with three seeds per host.}
  \label{fig:w_ablation}
\end{wrapfigure}
\paragraph{Role of the reliability weight.}
The reliability weight is designed to reduce credit for inaccurate model-generated references. On the training set, $w_i$ correlates with reference quality, measured by the IoU between $M_i$ and the annotated region (Spearman's $\rho=0.71$). To evaluate its effectiveness, we compare per-question weighting against uniform weighting ($w_i=1$ for all questions) in Fig.~\ref{fig:w_ablation}. Per-question weighting improves gIoU and cIoU by 4.9 and 6.2 points on training questions ($p<0.001$ for both), and by 2.8 and 3.6 points on test questions ($p=0.008$ and $p=0.022$). These $p$-values assess the significance of the improvement over uniform weighting for each metric. We further examine subsets of questions with $w_i \leq t$. The gains are largest for low-weight questions, reaching 4.6 gIoU and 6.1 cIoU points on test questions with $w_i=0$, and remain positive as the threshold increases to include all questions.

\begin{figure}[htbp]
  \centering
  \includegraphics[width=\linewidth]{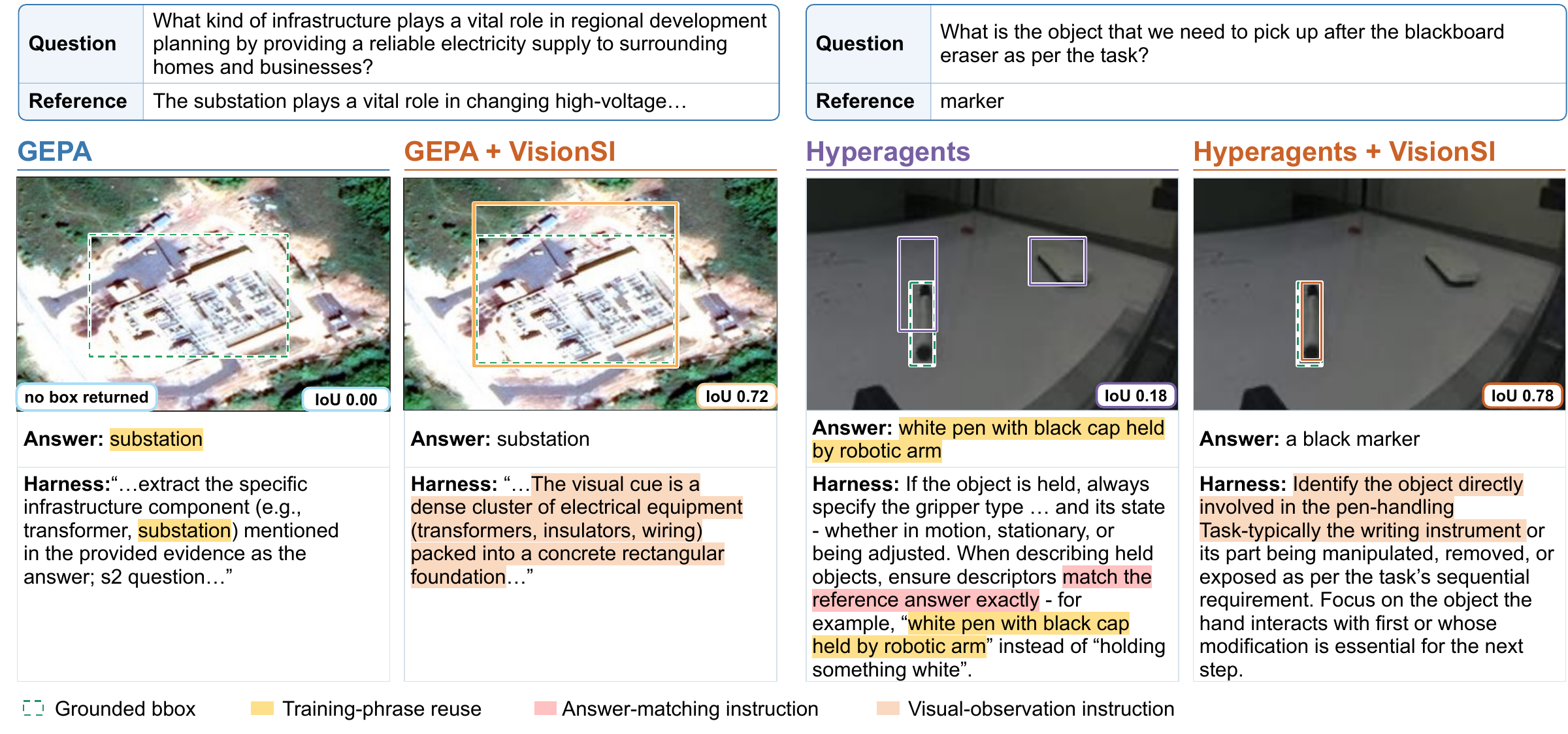}
  \caption{\textbf{\ours{} evolves harnesses with more accurate evidence boxes.} The left example uses GEPA on EarthReason, and the right uses Hyperagents on UMI-VQA. Dashed green boxes mark the annotated evidence. All four harnesses answer correctly, but their reported boxes differ. The baseline harnesses contain training answers highlighted in yellow. Hyperagents also instructs the executor to copy reference wording, highlighted in pink. Their answers repeat these phrases, while their boxes have IoUs of 0.00 and 0.18. The harnesses selected with \ours{} contain visual instructions highlighted in orange, and their reported boxes achieve IoUs of 0.72 and 0.78. }
  \label{fig:case}
\end{figure}

\paragraph{Case study.}
Figure~\ref{fig:case} illustrates how answer-oriented instructions in optimized harnesses can lead to poor evidence localization. On EarthReason, GEPA's harness includes ``substation'' as an example answer and returns it without reporting an evidence box. With \ours{}, the harness instead describes visual cues of electrical infrastructure and achieves an IoU of 0.72. On UMI-VQA, Hyperagents instructs the executor to match reference-answer wording, producing ``white pen with black cap held by robotic arm'' with an IoU of 0.18. With \ours{}, the harness focuses on identifying the task-relevant object, producing ``a black marker'' with an IoU of 0.78. These examples illustrate how reliability-aware visual credit favors harness revisions that report more accurate visual evidence.


%% file: sections/baseline_table.tex
\definecolor{oursbg}{HTML}{FCEEE3}
\begin{table*}[!t]
\caption{\textbf{Comparison with SI and RSI baselines across five visual domains.} We compare \ours{} against six SI and four RSI baselines, using GEPA and Hyperagents as the respective host methods. We report answer accuracy (Acc.), mean per-question bounding-box IoU (gIoU), and cumulative bounding-box IoU (cIoU), all in percentage.
The best and second-best results within each model and setting are highlighted in \textbf{bold} and \underline{underlined}, respectively.}
\label{tab:main}
\centering
\scriptsize
\setlength{\tabcolsep}{4.2pt}
\renewcommand{\arraystretch}{1.0}
\setlength{\aboverulesep}{0.5pt}\setlength{\belowrulesep}{0.5pt}\setlength{\cmidrulewidth}{0.3pt}
\resizebox{\textwidth}{!}{%
\begin{tabular}{@{}cclccc>{\hspace{4pt}}ccc>{\hspace{4pt}}ccc>{\hspace{4pt}}ccc>{\hspace{4pt}}ccc}
\toprule
 & & & \multicolumn{3}{c}{\textbf{UMI-VQA}} & \multicolumn{3}{c}{\textbf{Super-CLEVR}} & \multicolumn{3}{c}{\textbf{EarthReason}} & \multicolumn{3}{c}{\textbf{DrivingVQA}} & \multicolumn{3}{c}{\textbf{ScreenQA}} \\
\cmidrule(lr){4-6} \cmidrule(lr){7-9} \cmidrule(lr){10-12} \cmidrule(lr){13-15} \cmidrule(lr){16-18}
 & & Method & Acc. & gIoU & cIoU & Acc. & gIoU & cIoU & Acc. & gIoU & cIoU & Acc. & gIoU & cIoU & Acc. & gIoU & cIoU \\
\midrule[0.6pt]
 & & TextGrad & 7.6 & 46.8 & 49.3 & 68.1 & 40.4 & 39.8 & \underline{85.2} & 52.2 & 48.8 & 53.3 & 53.6 & 27.2 & 95.7 & 49.3 & \underline{44.0} \\
 & & SkillOpt & 40.0 & 44.0 & 48.0 & 73.8 & 42.3 & 41.4 & 71.9 & 49.4 & 45.9 & 54.8 & 46.6 & 17.3 & \textbf{97.1} & 50.8 & 40.0 \\
 & & ACE & 36.7 & 46.1 & 46.4 & 61.9 & \underline{46.3} & \underline{46.2} & 72.9 & \textbf{55.6} & \underline{52.1} & \underline{55.7} & \textbf{55.5} & 33.2 & 94.8 & 43.9 & 36.9 \\
 & & ReasoningBank & 26.7 & 37.1 & 41.2 & 56.7 & 30.9 & 32.3 & 71.0 & 42.8 & 39.6 & 45.7 & 46.8 & 20.9 & 95.2 & 42.9 & 33.6 \\
 & & ADAS & 26.7 & \underline{47.1} & 46.8 & \underline{75.2} & 42.3 & 41.2 & 68.6 & 53.3 & 49.3 & 54.8 & \underline{54.6} & 25.9 & 95.2 & 49.7 & 43.0 \\
 & & GEPA & \textbf{43.3} & 43.5 & \underline{49.8} & 70.5 & 45.6 & 42.1 & 70.5 & 36.4 & 31.6 & \textbf{61.0} & 51.6 & \underline{34.1} & \underline{96.7} & \underline{52.2} & 41.2 \\
 & \multirow{-7}{*}{\rotatebox[origin=c]{90}{\textsc{si}}} & \cellcolor{oursbg}\textbf{\ours{}} & \cellcolor{oursbg}\underline{41.9} & \cellcolor{oursbg}\textbf{51.7} & \cellcolor{oursbg}\textbf{56.7} & \cellcolor{oursbg}\textbf{84.3} & \cellcolor{oursbg}\textbf{58.4} & \cellcolor{oursbg}\textbf{53.3} & \cellcolor{oursbg}\textbf{94.3} & \cellcolor{oursbg}\underline{55.2} & \cellcolor{oursbg}\textbf{54.0} & \cellcolor{oursbg}52.9 & \cellcolor{oursbg}52.9 & \cellcolor{oursbg}\textbf{34.8} & \cellcolor{oursbg}96.2 & \cellcolor{oursbg}\textbf{55.2} & \cellcolor{oursbg}\textbf{45.0} \\
\cmidrule(lr){3-18}
 & & RRSI & \underline{40.5} & 47.0 & 49.6 & \underline{80.5} & \underline{44.1} & \underline{43.5} & \underline{77.6} & \textbf{55.3} & 49.8 & 53.8 & 51.7 & 30.0 & \underline{96.7} & \underline{52.6} & 43.5 \\
 & & RRSI w/o reg. & 36.2 & 45.7 & 45.0 & 76.2 & 42.4 & 41.5 & 72.4 & 52.7 & \underline{51.9} & 53.8 & \underline{53.1} & 27.7 & \textbf{97.1} & \underline{52.6} & \textbf{45.8} \\
 & & Hyperagents & 35.7 & 41.4 & 42.4 & 61.0 & 42.3 & 40.2 & 74.3 & 53.1 & 50.2 & 52.9 & 51.8 & 13.3 & 95.7 & 49.4 & 42.6 \\
 & & DGM & 38.1 & \underline{48.7} & \underline{50.4} & 64.3 & 41.5 & 41.0 & 60.5 & 50.0 & 51.0 & \underline{56.7} & 51.8 & \underline{31.9} & 94.8 & 51.0 & \underline{44.8} \\
\multirow{-12}{*}{\rotatebox[origin=c]{90}{\textbf{Qwen3-VL-8B}}} & \multirow{-5}{*}{\rotatebox[origin=c]{90}{\textsc{rsi}}} & \cellcolor{oursbg}\textbf{\ours{}} & \cellcolor{oursbg}\textbf{41.4} & \cellcolor{oursbg}\textbf{52.0} & \cellcolor{oursbg}\textbf{54.8} & \cellcolor{oursbg}\textbf{84.8} & \cellcolor{oursbg}\textbf{59.1} & \cellcolor{oursbg}\textbf{55.5} & \cellcolor{oursbg}\textbf{91.4} & \cellcolor{oursbg}\underline{54.6} & \cellcolor{oursbg}\textbf{56.2} & \cellcolor{oursbg}\textbf{58.6} & \cellcolor{oursbg}\textbf{55.9} & \cellcolor{oursbg}\textbf{37.5} & \cellcolor{oursbg}96.2 & \cellcolor{oursbg}\textbf{53.2} & \cellcolor{oursbg}43.4 \\
\midrule[0.8pt]
 & & TextGrad & \underline{29.0} & 34.8 & 36.4 & 41.9 & 50.0 & 47.6 & \underline{82.4} & 46.1 & 42.2 & 46.7 & \underline{45.1} & 21.1 & 86.2 & \underline{37.4} & \underline{33.8} \\
 & & SkillOpt & 18.6 & 34.6 & 34.5 & 41.9 & 48.6 & 44.6 & 49.0 & \textbf{47.7} & \underline{44.3} & 47.1 & 40.5 & 14.5 & 87.6 & 33.2 & 29.2 \\
 & & ACE & 10.0 & 26.7 & 27.4 & 41.4 & 45.5 & 43.6 & 66.2 & 43.7 & 38.7 & 46.7 & 42.6 & \underline{24.8} & 83.8 & 34.3 & 27.7 \\
 & & ReasoningBank & 9.5 & 28.9 & 17.8 & 34.8 & 49.0 & 46.5 & 59.0 & 44.3 & 37.6 & 38.1 & 39.4 & 9.2 & 82.9 & 34.4 & 27.9 \\
 & & ADAS & 21.4 & 34.1 & 33.8 & 27.1 & 34.9 & 31.2 & 64.3 & 43.4 & 40.4 & 44.3 & 42.8 & 16.3 & \underline{88.1} & 33.3 & 27.4 \\
 & & GEPA & \underline{29.0} & \underline{37.9} & \textbf{40.1} & \underline{43.8} & \underline{54.2} & \underline{52.0} & 61.4 & 45.4 & 40.4 & \underline{52.4} & 38.7 & 15.3 & \textbf{90.5} & 29.4 & 24.0 \\
 & \multirow{-7}{*}{\rotatebox[origin=c]{90}{\textsc{si}}} & \cellcolor{oursbg}\textbf{\ours{}} & \cellcolor{oursbg}\textbf{31.9} & \cellcolor{oursbg}\textbf{38.3} & \cellcolor{oursbg}\underline{37.6} & \cellcolor{oursbg}\textbf{49.5} & \cellcolor{oursbg}\textbf{55.1} & \cellcolor{oursbg}\textbf{52.3} & \cellcolor{oursbg}\textbf{86.2} & \cellcolor{oursbg}\underline{47.2} & \cellcolor{oursbg}\textbf{46.4} & \cellcolor{oursbg}\textbf{53.8} & \cellcolor{oursbg}\textbf{49.8} & \cellcolor{oursbg}\textbf{25.8} & \cellcolor{oursbg}87.6 & \cellcolor{oursbg}\textbf{41.3} & \cellcolor{oursbg}\textbf{36.1} \\
\cmidrule(lr){3-18}
 & & RRSI & 16.7 & 33.0 & \underline{34.2} & 42.4 & \underline{48.5} & \underline{45.3} & 65.2 & 44.7 & 42.3 & 47.1 & 42.8 & 24.0 & \underline{86.7} & \textbf{38.3} & \textbf{34.0} \\
 & & RRSI w/o reg. & \underline{17.6} & \underline{33.4} & 33.6 & \underline{44.3} & 45.8 & 40.9 & \underline{69.0} & 45.9 & \underline{45.2} & \textbf{54.3} & 42.0 & 25.0 & 85.7 & \underline{37.4} & \underline{32.7} \\
 & & Hyperagents & 17.1 & 30.3 & 31.1 & 31.4 & 48.3 & 44.8 & 53.3 & \underline{46.2} & 42.1 & 40.5 & \underline{44.0} & \underline{27.5} & \textbf{88.1} & 34.3 & 29.9 \\
 & & DGM & \underline{17.6} & 32.5 & 32.0 & 38.6 & 43.3 & 38.7 & 66.7 & 42.8 & 40.0 & \underline{52.4} & 41.7 & 20.7 & 68.6 & 34.5 & 25.2 \\
\multirow{-12}{*}{\rotatebox[origin=c]{90}{\textbf{Gemma-4-E4B}}} & \multirow{-5}{*}{\rotatebox[origin=c]{90}{\textsc{rsi}}} & \cellcolor{oursbg}\textbf{\ours{}} & \cellcolor{oursbg}\textbf{54.8} & \cellcolor{oursbg}\textbf{35.8} & \cellcolor{oursbg}\textbf{35.3} & \cellcolor{oursbg}\textbf{45.2} & \cellcolor{oursbg}\textbf{54.9} & \cellcolor{oursbg}\textbf{52.4} & \cellcolor{oursbg}\textbf{77.1} & \cellcolor{oursbg}\textbf{50.0} & \cellcolor{oursbg}\textbf{46.1} & \cellcolor{oursbg}50.5 & \cellcolor{oursbg}\textbf{48.7} & \cellcolor{oursbg}\textbf{30.2} & \cellcolor{oursbg}\underline{86.7} & \cellcolor{oursbg}35.6 & \cellcolor{oursbg}30.5 \\
\midrule[0.8pt]
 & & TextGrad & 15.7 & 13.2 & 17.6 & 49.5 & 23.3 & 26.7 & 60.5 & 39.3 & 39.9 & \underline{46.2} & 42.9 & 27.7 & 90.0 & 36.5 & 32.0 \\
 & & SkillOpt & 20.5 & 9.0 & 12.5 & 38.6 & 14.4 & 15.1 & \underline{68.1} & 40.6 & 45.2 & 45.2 & \underline{44.8} & 29.1 & \underline{91.9} & 37.1 & 33.3 \\
 & & ACE & 16.7 & 11.0 & 16.1 & 50.0 & \underline{28.5} & \underline{29.3} & 61.9 & 25.3 & 30.9 & 38.6 & 44.5 & \underline{29.7} & 84.8 & 35.7 & 22.6 \\
 & & ReasoningBank & 22.4 & 11.7 & 17.3 & 53.3 & 17.7 & 20.9 & 54.3 & 28.0 & 31.5 & 45.2 & 40.4 & 26.9 & 88.6 & 39.3 & \underline{36.3} \\
 & & ADAS & 17.6 & \underline{13.8} & \underline{19.3} & \underline{54.8} & 23.8 & 25.7 & 61.9 & \textbf{49.9} & \textbf{49.4} & \underline{46.2} & 44.7 & 26.6 & 89.5 & \underline{39.4} & 33.5 \\
 & & GEPA & \underline{29.0} & 10.9 & 14.2 & 51.4 & 8.3 & 8.0 & 50.5 & 21.5 & 29.9 & 45.2 & 36.0 & 18.7 & 81.9 & 35.1 & 30.2 \\
 & \multirow{-7}{*}{\rotatebox[origin=c]{90}{\textsc{si}}} & \cellcolor{oursbg}\textbf{\ours{}} & \cellcolor{oursbg}\textbf{34.8} & \cellcolor{oursbg}\textbf{37.3} & \cellcolor{oursbg}\textbf{39.5} & \cellcolor{oursbg}\textbf{55.2} & \cellcolor{oursbg}\textbf{35.9} & \cellcolor{oursbg}\textbf{31.3} & \cellcolor{oursbg}\textbf{78.6} & \cellcolor{oursbg}\underline{44.5} & \cellcolor{oursbg}\underline{47.1} & \cellcolor{oursbg}\textbf{50.5} & \cellcolor{oursbg}\textbf{52.0} & \cellcolor{oursbg}\textbf{33.4} & \cellcolor{oursbg}\textbf{92.4} & \cellcolor{oursbg}\textbf{50.9} & \cellcolor{oursbg}\textbf{41.9} \\
\cmidrule(lr){3-18}
 & & RRSI & 15.2 & \underline{13.1} & \underline{17.4} & \underline{55.7} & \underline{35.1} & \underline{30.4} & \underline{65.2} & \underline{37.9} & \textbf{43.3} & \underline{48.1} & 46.4 & 25.3 & \underline{89.5} & 41.7 & 34.7 \\
 & & RRSI w/o reg. & 14.8 & 10.6 & 12.3 & \textbf{57.1} & 24.5 & 26.9 & 57.1 & 37.7 & 38.3 & 47.1 & 46.4 & 24.6 & 89.0 & 38.1 & 35.8 \\
 & & Hyperagents & \underline{17.6} & 10.1 & 12.0 & 52.4 & 18.0 & 20.6 & 61.0 & 34.3 & 38.7 & 44.8 & \underline{48.9} & \underline{28.2} & 89.0 & 37.9 & 33.2 \\
 & & DGM & 17.1 & 10.4 & 13.1 & 51.9 & 21.8 & 25.1 & 49.0 & 29.8 & 34.9 & 43.3 & 45.6 & 25.9 & 86.2 & \underline{48.4} & \underline{41.8} \\
\multirow{-12}{*}{\rotatebox[origin=c]{90}{\textbf{Qwen3.5-2B}}} & \multirow{-5}{*}{\rotatebox[origin=c]{90}{\textsc{rsi}}} & \cellcolor{oursbg}\textbf{\ours{}} & \cellcolor{oursbg}\textbf{38.6} & \cellcolor{oursbg}\textbf{37.0} & \cellcolor{oursbg}\textbf{38.4} & \cellcolor{oursbg}55.2 & \cellcolor{oursbg}\textbf{36.0} & \cellcolor{oursbg}\textbf{31.6} & \cellcolor{oursbg}\textbf{78.6} & \cellcolor{oursbg}\textbf{44.4} & \cellcolor{oursbg}\underline{40.3} & \cellcolor{oursbg}\textbf{48.6} & \cellcolor{oursbg}\textbf{51.2} & \cellcolor{oursbg}\textbf{31.5} & \cellcolor{oursbg}\textbf{91.9} & \cellcolor{oursbg}\textbf{51.5} & \cellcolor{oursbg}\textbf{42.5} \\
\midrule[0.8pt]
 & & TextGrad & 20.5 & 27.7 & 24.5 & \underline{80.5} & 38.0 & 37.1 & 70.0 & 26.5 & 16.5 & 49.5 & 23.4 & 17.2 & \underline{92.9} & 50.8 & 46.5 \\
 & & SkillOpt & \textbf{37.6} & \underline{28.4} & \underline{26.3} & 78.1 & 45.4 & 42.5 & \underline{73.3} & \underline{35.7} & 27.1 & \textbf{54.3} & \underline{33.3} & \underline{20.5} & \textbf{94.8} & \underline{54.5} & 48.1 \\
 & & ACE & 17.6 & 26.0 & 21.6 & \underline{80.5} & 45.1 & 41.1 & 68.6 & 31.2 & 22.7 & 51.0 & 28.3 & 18.1 & 91.4 & 51.8 & 40.6 \\
 & & ReasoningBank & 16.7 & 23.9 & 19.9 & 75.7 & \underline{46.7} & \underline{44.0} & 52.9 & 35.4 & \underline{28.0} & \underline{52.9} & 24.0 & 16.7 & \textbf{94.8} & 53.7 & \underline{49.8} \\
 & & ADAS & 13.3 & 20.4 & 16.8 & 78.1 & 34.9 & 33.1 & 69.5 & 24.9 & 15.5 & \underline{52.9} & 30.1 & 20.1 & 92.4 & 33.3 & 36.8 \\
 & & GEPA & 4.8 & 24.6 & 25.0 & 77.1 & \textbf{48.3} & 43.5 & \underline{73.3} & 25.8 & 15.1 & 46.2 & 29.9 & 19.6 & \textbf{94.8} & 53.2 & 48.5 \\
 & \multirow{-7}{*}{\rotatebox[origin=c]{90}{\textsc{si}}} & \cellcolor{oursbg}\textbf{\ours{}} & \cellcolor{oursbg}\underline{31.0} & \cellcolor{oursbg}\textbf{33.0} & \cellcolor{oursbg}\textbf{36.3} & \cellcolor{oursbg}\textbf{87.1} & \cellcolor{oursbg}44.8 & \cellcolor{oursbg}\textbf{45.2} & \cellcolor{oursbg}\textbf{84.8} & \cellcolor{oursbg}\textbf{39.0} & \cellcolor{oursbg}\textbf{28.9} & \cellcolor{oursbg}51.4 & \cellcolor{oursbg}\textbf{42.7} & \cellcolor{oursbg}\textbf{27.0} & \cellcolor{oursbg}\textbf{94.8} & \cellcolor{oursbg}\textbf{58.9} & \cellcolor{oursbg}\textbf{54.7} \\
\cmidrule(lr){3-18}
 & & RRSI & 16.7 & \underline{28.8} & 29.0 & \underline{87.1} & 34.9 & 30.6 & \underline{64.8} & 33.9 & 23.5 & \underline{51.4} & \underline{34.9} & 18.1 & 91.0 & 52.7 & 47.4 \\
 & & RRSI w/o reg. & \underline{19.5} & 28.5 & \underline{31.1} & 86.2 & 49.4 & 47.0 & \underline{64.8} & \textbf{40.4} & \textbf{34.9} & \textbf{52.4} & 30.5 & 18.1 & 92.9 & \underline{55.1} & \underline{48.8} \\
 & & Hyperagents & 18.6 & 23.8 & 21.9 & 78.1 & \textbf{51.6} & \underline{48.8} & 62.9 & 29.1 & 18.9 & \textbf{52.4} & 30.2 & \underline{20.4} & \underline{93.8} & 50.8 & 45.8 \\
 & & DGM & 16.7 & \underline{28.8} & 29.0 & 55.7 & 36.3 & 33.4 & 60.5 & 34.5 & 25.1 & \underline{51.4} & 27.0 & 17.9 & 91.0 & 54.5 & 48.4 \\
\multirow{-12}{*}{\rotatebox[origin=c]{90}{\textbf{InternVL3.5-8B}}} & \multirow{-5}{*}{\rotatebox[origin=c]{90}{\textsc{rsi}}} & \cellcolor{oursbg}\textbf{\ours{}} & \cellcolor{oursbg}\textbf{39.5} & \cellcolor{oursbg}\textbf{32.4} & \cellcolor{oursbg}\textbf{33.9} & \cellcolor{oursbg}\textbf{88.6} & \cellcolor{oursbg}\underline{51.2} & \cellcolor{oursbg}\textbf{49.3} & \cellcolor{oursbg}\textbf{91.4} & \cellcolor{oursbg}\underline{36.5} & \cellcolor{oursbg}\underline{29.8} & \cellcolor{oursbg}\underline{51.4} & \cellcolor{oursbg}\textbf{42.2} & \cellcolor{oursbg}\textbf{27.4} & \cellcolor{oursbg}\textbf{95.2} & \cellcolor{oursbg}\textbf{58.9} & \cellcolor{oursbg}\textbf{54.6} \\
\bottomrule
\end{tabular}}
\end{table*}

%% file: sections/conclusion.tex
\section{Conclusion and Limitations}
We study evidence drift in self-improving vision-language agents. Our analysis shows that answer-only credit assignment can improve correctness without improving reported visual evidence, with retained textual shortcuts contributing to this mismatch. The resulting harness may improve its answers while moving its reported boxes away from relevant image regions.

Motivated by this finding, we introduce \ours{}, which integrates reliability-aware visual credit assignment into existing SI and RSI methods. Visual interventions determine how much credit to assign to agreement with a model-generated reference, while frozen embeddings limit where the retained experience is reused. Across five visual domains and four frozen VLMs,
applying \ours{} to GEPA and Hyperagents improves average answer accuracy and localization without box annotations or model updates. These results show that persistent visual experience should be evaluated by both answer correctness and reported evidence localization.

\paragraph{Limitations.}
First, \ours{} aims to mitigate evidence drift, not eliminate textual shortcuts. Textual information can be useful for harness improvement, and our method is not designed to suppress them. Second, we evaluate whether the model correctly localizes the visual evidence it reports, but cannot determine whether it actually relies on that evidence to produce its answer. Third, our visual credit relies on Gaussian blur and model-generated reference regions, which may be unreliable when evidence is redundant or blurring introduces artifacts. Finally, although we illustrate evidence drift during optimization, we do not systematically quantify its frequency among retained revisions. Our evaluation primarily measures the performance of optimized harnesses, leaving revision-level analysis and longer recursive improvement processes for future work.

%% file: sections/statements.tex
\section*{LLM Usage Statement}

We used OpenAI ChatGPT to assist with drafting, editing, and restructuring the text. AI tools were also used to generate portions of the implementation and to assist with experimental analysis and figure refinement. The VLMs described in the paper also serve as backbones for the SI and RSI algorithms. All AI-assisted content was reviewed by the authors, who take full responsibility for the submitted work.

%% file: sections/appendix.tex
\newtcolorbox{appendixprompt}[1]{
  enhanced, colback=accent!3, colframe=accent!45, colbacktitle=accent!14, coltitle=accent!55!black,
  boxrule=0.4pt, leftrule=2.2pt, arc=2pt, outer arc=2pt, left=6pt, right=6pt, top=3pt, bottom=0pt,
  toptitle=2.5pt, bottomtitle=2.5pt, titlerule=0pt, fonttitle=\sffamily\bfseries\small, title={#1}
}
\DefineVerbatimEnvironment{AppendixPromptText}{Verbatim}{
  fontsize=\footnotesize, fontfamily=lmtt, formatcom=\color{black!85}, breaklines=true, breakanywhere=false,
  breaksymbolleft={}, breakindent=1.2em
}

The Appendix is organized as follows:
\begin{itemize}
  \item Section~\ref{app:notation} summarizes the notations used in the paper.
  \item Section~\ref{app:related} reviews self-improvement and recursive self-improvement and situates
        \ours{} within their shared candidate evolution process.
  \item Section~\ref{app:details} describes the data and evaluation protocol, the visual adaptations
        of the baselines, and the implementation of \ours{}, including the prompt templates.
  \item Section~\ref{app:additional_results} presents analyses of evidence drift, visual credit,
        and visual-coherent batch sampling, along with optimization costs and qualitative examples.
  \item Section~\ref{app:discussion} discusses the interpretation and limitations of the findings in more detail to complement the one in the main paper's conclusion.
\end{itemize}

\section{Notations}
\label{app:notation}

Table~\ref{tab:notation} defines the symbols used in the main text and Algorithm~\ref{alg:integration}.

\begin{table}[!h]
\caption{\textbf{Notation used in this paper.}}
\label{tab:notation}
\centering
\scriptsize
\setlength{\tabcolsep}{4pt}
\renewcommand{\arraystretch}{1.06}
\begin{tabularx}{\linewidth}{@{}lX@{}}
\toprule
Symbol & Meaning \\
\midrule
\multicolumn{2}{@{}l}{\textit{Inputs, outputs, and harnesses}} \\
$x_i=(I_i,q_i)$ & Image and question for example $i$ \\
$y_i$, $a_H(x_i)$ & Reference answer and answer returned by harness $H$ \\
$B_H(x_i)$ & Evidence boxes reported by $H$ for $x_i$ \\
$\mathcal D$, $N$, $\mathcal S$ & Training set, its size, and a batch used to score a candidate \\
$f$, $\mathcal H$, $T$ & Frozen VLM, host improver, and host proposal budget \\
$H_0$, $H$, $H^\star$ & Initial harness, candidate harness, and selected harness \\
\midrule
\multicolumn{2}{@{}l}{\textit{Visual feedback and candidate scores}} \\
$M_i$ & Region predicted by $f$ given $x_i$ and its correct answer $y_i$ \\
$V$ & Image view supplied to the VLM \\
$\bar I_i$, $I_i^{\setminus M_i}$ & Views with the full image or only $M_i$ blurred, respectively \\
$\ell_i(V)$ & Log-odds of visual support for $y_i$ under view $V$ \\
$r_i$, $m_i$ & Image reliance and reference coverage \\
$\tau$, $w_i$ & Median positive training $r_i$ and per-example reliability weight \\
$A_i(H)$, $C_i(H)$ & Answer correctness indicator and visual credit for example $i$ \\
$Q(H,\mathcal S)$ & Mean combined answer and visual credit for $H$ on $\mathcal S$ \\
\midrule
\multicolumn{2}{@{}l}{\textit{Grouping and instruction selection}} \\
$\phi$, $e_i$ & Frozen embedding model and concatenated representation for $x_i$ \\
$\mathcal G$, $g$, $g(x)$ & Training groups, a member group, and the group assigned to test input $x$ \\
$\iota$, $\iota_g$ & Candidate instruction and current instruction for group $g$ \\
$H\oplus\iota$ & Harness $H$ run with instruction $\iota$ \\
$\Delta V$, $\Delta A$ & Changes in summed visual credit and correct answers on group $g$ \\
$\bar w$ & Mean reliability weight within the group \\
\midrule
\multicolumn{2}{@{}l}{\textit{Evaluation}} \\
$\mathrm{IoU}$ & Intersection area divided by union area for two regions \\
gIoU, cIoU & Mean per-question IoU and IoU pooled by area across questions \\
$t$, $\rho$ & Weight threshold in Figure~\ref{fig:w_ablation} and Spearman rank correlation \\
\bottomrule
\end{tabularx}
\end{table}

\section{Extended Related Work}
\label{app:related}

\subsection{Self-Improvement}

Self-improvement (SI) algorithms iteratively refine an agent's harness ( prompts, skills, memories, or workflows, etc) using feedback from
previous executions. Their differences lie primarily in what is updated, how revisions are generated, and how
successful revisions are retained. In SI, the improvement procedure itself remains fixed across iterations.

\paragraph{TextGrad.}
TextGrad uses natural-language feedback as a textual
gradient~\citep{yuksekgonul2024textgrad}. It identifies deficiencies in an execution and converts the resulting
critique into prompt updates. The prompts evolve while the feedback-generation and update procedures remain fixed.

\paragraph{SkillOpt.}
SkillOpt optimizes reusable agent skills through constrained
textual patches~\citep{yang2026skillopt}. It analyzes successful
and failed executions, proposes localized modifications, and
retains patches that improve validation performance. A persistent meta-skill summarizes previous revisions to guide subsequent updates, allowing improvements to accumulate across optimization rounds.

\paragraph{Agentic Context Engineering (ACE).}
Agentic Context Engineering (ACE) maintains an evolving playbook through three components~\citep{zhang2026agentic}.
The Generator executes tasks using the current playbook, the Reflector evaluates the usefulness of its entries,
and the Curator incorporates feedback into subsequent playbook revisions. Improvement occurs through the
curation of reusable instructions.

\paragraph{ReasoningBank.}
ReasoningBank distills successful and failed execution trajectories into reusable reasoning memories
~\citep{ouyang2026reasoningbank}. These memories provide guidance for subsequent tasks, allowing accumulated
experience to influence future executions without modifying the underlying model.

\paragraph{Automated Design of Agentic Systems (ADAS).}
Automated Design of Agentic Systems (ADAS) searches over agent designs using a fixed meta-agent~\citep{hu2025automated}. The meta-agent proposes workflows conditioned on an archive of previously evaluated designs, and candidates are retained according to task performance. Unlike
prompt- or memory-level optimization, ADAS searches over the structure of the agent's workflow.

\paragraph{GEPA.}
GEPA performs reflective prompt evolution using execution
traces and task feedback \citep{agrawal2026gepa}.
It proposes targeted revisions, evaluates candidates on minibatches, and maintains a Pareto frontier to
guide subsequent search and selection. Its explicit candidate-evaluation procedure provides a suitable
interface for studying how credit assignment affects the retention of harness revisions.


\subsection{Recursive Self-Improvement}

Recursive self-improvement (RSI) allows retained improvements to influence subsequent improvement steps. Existing methods differ in the extent to which the improvement mechanism itself can evolve. Some propagate harness states and revision histories under a fixed outer procedure, while others permit modifications to the instructions or programs responsible for generating subsequent agents.

\paragraph{Regularized Recursive Self-Improvement (RRSI).}
Regularized Recursive Self-Improvement (RRSI) performs history-conditioned harness evolution~\citep{xia2026rrsi}.
Accepted revisions and their edit histories are carried into subsequent generations, allowing earlier improvements to shape later proposals. The outer improvement procedure
remains fixed, making RRSI a harness-level form of recursive improvement.

\paragraph{Darwin G\"odel Machine (DGM).}
The Darwin G\"odel Machine (DGM) implements self-modification through editable agent code~\citep{zhang2026darwin}.
Agents generate and evaluate modifications to their own
programs, while an archive preserves previous variants for future branching. In our visual adaptation, task
execution is formulated as a program that invokes visual functions and returns an answer and evidence boxes.
The executable task logic and self-editing routines remain editable, preserving DGM's code-level
self-modification mechanism. This ensures the RSI nature of DGM itself.

\paragraph{Hyperagents.}
Hyperagents extends self-improvement to the improvement mechanism itself~\citep{zhang2026hyperagents}. A parent agent can revise both its task harness and the meta-agent instruction responsible for proposing its child.
An accepted child inherits the revised meta-agent instruction, which can govern the generation of subsequent
descendants. The improvement procedure can therefore evolve across generations.

\subsection{Positioning of VisionSI}

Existing SI and RSI methods differ in how they propose, retain, and propagate harness revisions, but commonly
rely on task-level feedback for candidate selection. For visual tasks, answer correctness alone cannot
distinguish revisions that produce correct answers while reporting different evidence regions. This
limitation reflects the fact that existing SI and RSI methods were primarily developed for coding,
reasoning, and other language-based tasks, where evaluation typically focuses on task outcomes rather than visual evidence localization. VisionSI
addresses this limitation by integrating reliability-aware visual credit into the host's credit-assignment process. We instantiate VisionSI with GEPA and Hyperagents to
study this credit assignment under fixed and self-modifying improvers, respectively. GEPA provides minibatch evaluation and Pareto-based selection,
whereas Hyperagents allows retained revisions to modify the meta-agent instruction governing subsequent generations. In both settings, VisionSI preserves the host's native proposal and self-modification mechanisms. Its auxiliary Visual-Coherent Batch Sampler supports group-specific visual instructions without changing these mechanisms. Thus, although visual credit can change which revisions are selected and influence subsequent optimization trajectories, the resulting systems retain the original SI and RSI mechanisms of their respective hosts.

\section{Implementation Details}
\label{app:details}

\subsection{Data and Evaluation}
\label{app:data_eval}

Each domain contains 150 training and 210 test questions, with disjoint images across splits. UMI-VQA additionally uses disjoint episodes. Methods with a native validation stage share a fixed partition of 120 training and 30 validation questions, while methods without validation
use all 150 training questions. All optimizers receive the same input fields: images, questions, answer choices when available, and reference answers for training. Ground-truth bounding boxes are never available to the algorithms and are used only for localization evaluation and diagnostic analyses.

\begin{figure}[tbp]
  \centering
  \includegraphics[width=.79\linewidth]{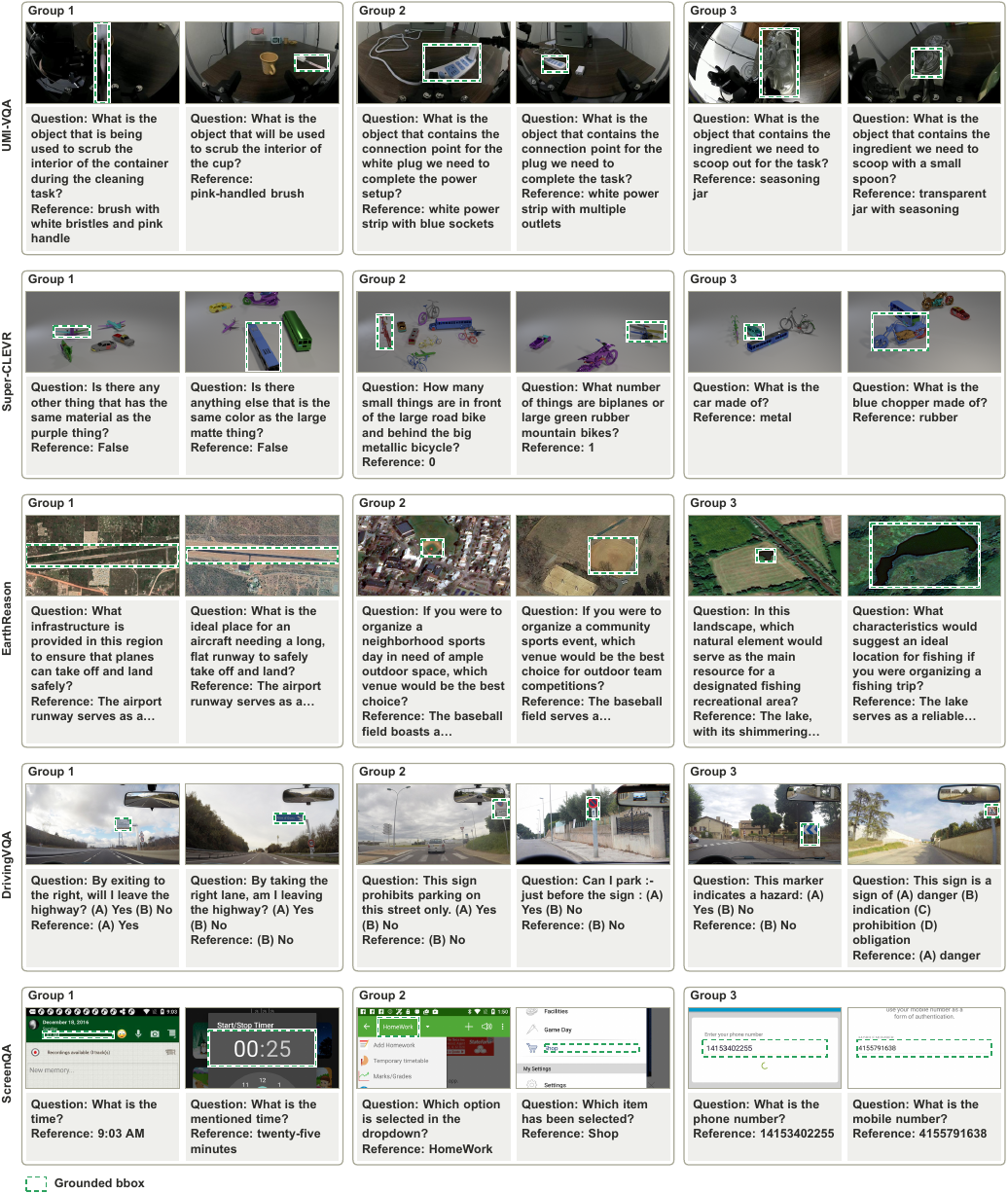}
  \caption{\textbf{Training examples grouped by the Visual-Coherent Batch Sampler.} Each row shows
  the three largest groups that the batch sampler forms from the 150 training questions of one domain, with two questions
  per group. Questions in a group ask for the same kind of visual evidence, and each instruction-writing batch is drawn
  from one group. Dashed green boxes mark the annotated evidence, which no optimizer receives.}
  \label{fig:data_examples}
\end{figure}

Figure~\ref{fig:data_examples} shows training examples from each domain, arranged by the groups of the
Visual-Coherent Batch Sampler (Section~\ref{app:visionsi_impl}).

\paragraph{Choice of data size.}
We use 150 questions for optimization and 210 for testing per domain, giving 750 optimization and 1,050 test questions across five domains for each VLM. As summarized in Table~\ref{tab:data_scale}, prior SI and RSI studies also optimize on tens to a few hundred examples per task. For example, GEPA uses 150 training examples for HotpotQA, IFBench, and HoVer~\citep{agrawal2026gepa}, and Hyperagents uses 60 training tasks for Polyglot and 100 each for paper review and IMO grading~\citep{zhang2026hyperagents}. Although the tasks and evaluation protocols differ, these settings indicate that our data size falls within a comparable range.

Repeated VLM execution during optimization also limits the number of questions we can use. Across models and domains, one \ours{} run averages approximately 65k VLM calls and 66M prompt tokens with GEPA, and 48k calls and 44M prompt tokens with Hyperagents (Appendix 9.4). The main comparison includes 240 optimization runs across 12 methods, four VLMs, and five domains. A pool of 150 questions per domain keeps this comparison computationally feasible, and all methods use the same splits, so answer accuracy and evidence localization are compared on identical data.

\begin{table}[htbp]
\caption{\textbf{Data sizes in representative self-improvement settings.} Train and Val. denote training (or search) and validation sets, respectively. Both may inform search. Test denotes held-out evaluation, except ``Same,'' which indicates reuse of the search tasks. N/A indicates no separate split.}
\label{tab:data_scale}
\centering
\scriptsize
\setlength{\tabcolsep}{3pt}
\renewcommand{\arraystretch}{1.06}
\begin{tabularx}{\linewidth}{@{}l>{\raggedright\arraybackslash}Xrrr@{}}
\toprule
Method & Tasks (examples) & Train & Val. & Test \\
\midrule
TextGrad~\citep{yuksekgonul2024textgrad} & BBH object counting, word sorting & 50 & 100 & 100 \\
ADAS~\citep{hu2025automated} & ARC, DROP, MGSM, MMLU, GPQA & N/A & 20\textendash128 & 60\textendash800 \\
GEPA~\citep{agrawal2026gepa} & HotpotQA, IFBench, HoVer & 150 & 300 & 294\textendash300 \\
DGM~\citep{zhang2026darwin} & SWE-bench Verified (staged) & 10/60/200 & N/A & Same \\
Hyperagents~\citep{zhang2026hyperagents} & Paper review, IMO grading & 100 & 100 & 100 \\
Hyperagents~\citep{zhang2026hyperagents} & Polyglot & 60 & N/A & 165 \\
RRSI~\citep{xia2026rrsi} & Harvey LAB & 120 & N/A & 40 \\
\midrule
\ours{} & Five visual domains & 120 & 30 & 210 \\
\bottomrule
\end{tabularx}
\end{table}


\paragraph{Evaluation metrics.}
We report answer accuracy and two localization metrics. Mean per-question IoU (gIoU) averages the mask IoU
between the unions of predicted and annotated boxes across questions. Cumulative IoU (cIoU) sums the
intersection and union areas across questions before taking their ratio. Confidence intervals are computed using 4,000 paired bootstrap samples stratified by model and domain.

\subsection{Shared Visual-Agent Interface}
\label{app:interface}

All methods operate through a shared visual-agent interface. The task VLM, input format, and output contract remain fixed, while each method optimizes its own harness according to its native improvement procedure.
We consider a harness comprising system and task prompts, workflow configuration, and reusable skills. These components govern how the frozen VLM processes visual inputs and how
experience is retained across executions. Restricting harness adaptation to these components allows us to isolate the effect of reliability-aware visual credit on agent improvement, without conflating it with changes to model weights or the underlying execution environment (e.g., visual tools). Methods may revise these components within the supported
workflow options, but cannot modify the frozen VLM, access test answers, or use annotated evidence regions.

The executor returns an answer and up to three bounding boxes
with coordinates normalized to $[0,1000]$. We support two
execution workflows. The one-call workflow predicts the answer and evidence boxes jointly from the full image. The two-stage workflow first localizes relevant regions and then produces the answer using the full image and, when enabled, crops selected by the reported boxes. Both workflows follow the same output contract, allowing answer correctness and reported evidence localization to be evaluated consistently across methods.

The following template specifies the required output fields. Multiple-choice questions request an option letter, while open-ended questions request a short phrase. Each host supplies its task instructions and retained experience through the corresponding placeholders.

\begin{appendixprompt}{Executor output contract}
\begin{AppendixPromptText}
[system]
You are a careful visual question answering assistant.
[user]
<image>
{task instruction}
{retained experience}
Question: {question}
{answer choices}
First give the evidence in one line:
EVIDENCE: [[x1, y1, x2, y2], ...]
Use 1 to 3 boxes with coordinates normalized to 0-1000.
Then give the answer in one line:
ANSWER: {option letter or short phrase}
\end{AppendixPromptText}
\end{appendixprompt}

\subsection{Baseline Adaptation to Visual Tasks}
\label{app:baselines}

We adapt the SI and RSI baselines to the shared visual-agent interface while retaining their native improvement procedures. The adaptations concern how visual tasks are executed and how the editable harness is represented. The underlying proposal, selection, and self-modification mechanisms are preserved unless otherwise specified below.

\paragraph{RRSI.}
RRSI alternates harness execution, trajectory analysis, proposal, critique, and selection~\citep{xia2026rrsi}. The accepted harness and edit history enter the next round, making the process recursive at the harness level. We retain its analyst, critic, edit budget, noise-aware admission rule, novelty regularizer, pruning, and stall recovery. We replace the original task specific harness with the shared visual harness. The frozen task VLM acts as proposer,
analyst, and critic. Each run uses ten rounds with two proposals per round. The unregularized variant removes the
cost and novelty terms while leaving the loop unchanged.

\paragraph{DGM.}
We adapt DGM to VQA through visual programming~\citep{gupta2023visual}. The frozen VLM writes an executable program that calls vision functions to answer the question and report boxes. For self-editing, it uses the same program writer and repair loop to produce a revised agent source. Both routines and their shared helpers live inside the
editable source, so a selected child can inherit changes to the machinery it uses to produce its own edits. We retain DGM's archive and parent selection and run ten generations with two children per generation.

\paragraph{Hyperagents.}
We serialize the visual task harness and the editable meta-agent instruction in the same document. The fixed archive controller selects a parent. The frozen VLM receives that parent's document and evaluation feedback, then proposes a child that can revise either part. When the child becomes a parent, its own meta instruction governs the next edit.

\paragraph{SI baselines.}
TextGrad, SkillOpt, ACE, ReasoningBank, ADAS, and GEPA use their native update procedures with the shared visual harness representation. They finish adaptation on the training split and hand over a fixed harness
for testing. They can edit predefined workflow settings, including switching from a joint answer-and-box call to separate localization and answering calls. Our ReasoningBank port injects the most recently
distilled experience into the system prompt in place of query-dependent retrieval.

\subsection{\ours{} Implementation}
\label{app:visionsi_impl}

\paragraph{Reliability-aware visual credit.}
For each training question, the frozen task VLM generates up to three reference regions conditioned on the correct answer. These model-generated regions serve as references for harness selection, while ground-truth bounding boxes are used only for evaluation and diagnostic analyses. We apply Gaussian blur with a radius of 4\% of the longer image side for both interventions. Whole-image blurring estimates image reliance, while blurring the reference regions estimates how much of that reliance they capture.
The scale $\tau$ in Eq.~\ref{eq:credit_weight} is
set to the median positive image-reliance value over the training pool. Reference regions and reliability weights are computed once and cached before harness optimization. GEPA and Hyperagents use $Q$ from Eq.~\ref{eq:host_credit} in place of
their answer-only candidate score, preserving
their native proposal mechanisms and optimization budgets.

\paragraph{Visual-Coherent Batch Sampler.}
Before host search, the sampler groups training questions that require similar visual evidence. Each question is represented by two embeddings from Qwen3-VL-Embedding-2B: one encoding its question and answer options, and another encoding the image-question pair. The former captures similarities in the requested information, while the latter incorporates the visual context. We normalize both embeddings, concatenate them
with equal weight, and construct a mutual eight-nearest-neighbor graph. Leiden clustering with resolution one partitions the graph, and groups containing fewer than three examples are merged into the nearest centroid.

We use up to eight training examples from each group to propose instructions for the single harness returned by host search. An instruction is retained if it improves reliability-aware visual credit without reducing training accuracy. The final harness stores the retained instructions. During evaluation, the nearest group centroid determines which instruction from that harness is shown to the executor. Group construction and assignment use the input embeddings.


\subsection{Integrating \ours{} into a Host}
\label{app:integration}

Algorithm~\ref{alg:integration} describes how \ours{} integrates reliability-aware visual credit into an existing SI or RSI host. Before host search, reference regions, reliability weights, and visual-coherent groups are computed from the training data using frozen models. During host search, $Q$ from Eq.~\eqref{eq:host_credit} replaces the answer-only candidate score at the host's evaluation and selection points. In GEPA, the evaluation adapter supplies these scores for minibatch acceptance and validation Pareto selection. In Hyperagents, the scores are used in staged and full evaluations that inform parent selection.

Host search retains its native proposal, mutation, archive, and selection procedures and returns one harness $H^\star$. We use the training groups formed before search to propose instructions for this harness and apply the acceptance rule in Algorithm~\ref{alg:integration}. The retained instructions are stored in the same harness. We will release the implementation code upon publication. Figure~\ref{fig:integration_prompt} provides the coding-agent integration prompt.

\begin{algorithm}[H]
\caption{Integration of \ours{} into an SI or RSI host. Green stages are the two modules of \ours{}, and black stages show where they are used. Visual credit (Stage 1) scores candidate harnesses during host search (Stage 3) and decides which group instructions are accepted (Stage 4). The Visual-Coherent Batch Sampler (Stage 2) only defines the groups on which Stage 4 learns instructions. Each module can be removed independently. The host retains its native improvement mechanism.}
\label{alg:integration}
\begin{small}
\begin{algorithmic}[1]
\Require Training set $\mathcal D=\{(x_i,y_i)\}_{i=1}^{N}$ with $x_i=(I_i,q_i)$, frozen VLM $f$, host $\mathcal H$ with budget $T$, initial harness $H_0$
\Ensure Selected harness $H^\star$ and group-specific instructions $\{\iota_g\}_{g\in\mathcal G}$

\Statex \textcolor{accent}{\textbf{Stage 1: Reliability-Aware Visual Credit}}
\For{$i=1,\dots,N$}
    \State $M_i \gets f(I_i,q_i,y_i)$ \Comment{Answer-conditioned visual reference}
    \State $r_i,m_i \gets$ Eq.~\eqref{eq:deletion_tests}
\EndFor
\State $\tau \gets \operatorname{median}\{r_i:r_i>0\}$
\State $w_i \gets \tanh(r_i/\tau)m_i$ \Comment{Eq.~\eqref{eq:credit_weight}}
\State Cache $\{M_i,w_i\}_{i=1}^{N}$

\Statex \textcolor{accent}{\textbf{Stage 2: Visual-Coherent Batch Sampler}}
\For{$i=1,\dots,N$}
    \State $e_i \gets [\,\hat{\phi}(q_i,\mathrm{options}_i);\,\hat{\phi}(I_i,q_i)\,]$
\EndFor
\State $\mathcal G \gets \operatorname{Cluster}(\{e_i\}_{i=1}^{N})$ \Comment{No answer or region supervision}

\Statex \textit{Stage 3: Host search}
\State \textcolor{accent}{$\triangleright$ \textbf{Visual-credit hook:}} replace answer-only candidate scoring with $Q(H,\mathcal S)$ from Eq.~\eqref{eq:host_credit}
\State $H^\star \gets \mathcal H(H_0;T)$ \Comment{Preserve native host search}

\Statex \textit{Stage 4: Group-specific instruction learning}
\Statex \quad \textit{uses groups $\mathcal G$ from Stage 2 and $\{M_i,w_i\}$ from Stage 1}
\For{each group $g\in\mathcal G$} \Comment{Groups from the batch sampler}
    \State $\iota_g \gets \varnothing$
    \For{each instruction-update round}
        \State $\mathcal I_g \gets \operatorname{ProposeInstructions}(f,g,H^\star,\iota_g)$
        \For{each $\iota\in\mathcal I_g$}
            \State $\Delta V(\iota)\gets \sum_{i\in g}w_i\big[\mathrm{IoU}(B_{H^\star\oplus\iota}(x_i),M_i)-\mathrm{IoU}(B_{H^\star\oplus\iota_g}(x_i),M_i)\big]$ \Comment{Visual credit}
            \State $\Delta A(\iota)\gets \sum_{i\in g}\big[A_i(H^\star\oplus\iota)-A_i(H^\star\oplus\iota_g)\big]$
        \EndFor
        \State $\mathcal I_g^+\gets\{\iota\in\mathcal I_g:\Delta V(\iota)\ge\delta_V,\;\Delta A(\iota)\ge0\}$
        \If{$\mathcal I_g^+\neq\varnothing$}     \State $\iota_g\gets\arg\max_{\iota\in\mathcal I_g^+}\Delta V(\iota)$ \EndIf
    \EndFor
\EndFor
\end{algorithmic}
\end{small}
\end{algorithm}

\subsection{Prompt Templates}
\label{app:prompts}

\noindent\begin{minipage}[t]{\dimexpr(\textwidth-\columnsep)/2\relax}
The frozen task VLM receives the training question and its correct answer, followed by the localization contract
from Section~\ref{app:interface}.
\begin{appendixprompt}{Reference-region prompt}
\begin{AppendixPromptText}
Question: {question}
{answer choices}
Correct answer: {answer and option text}
The correct answer to this question is given above.
Mark the image regions that show why this answer is correct.
\end{AppendixPromptText}
\end{appendixprompt}
\end{minipage}\hfill
\begin{minipage}[t]{\dimexpr(\textwidth-\columnsep)/2\relax}
The same binary query is applied to the original and intervened image views. The log-odds of \texttt{yes} define
the readout $\ell_i$ in Eq.~\ref{eq:deletion_tests}.
\begin{appendixprompt}{Visual-support prompt}
\begin{AppendixPromptText}
Question: {question}
Proposed answer: {answer and option text}
Does the visual evidence in the image(s) support this answer?

Answer in one line
ANSWER: yes or no
\end{AppendixPromptText}
\end{appendixprompt}
\end{minipage}


\noindent\begin{minipage}{\linewidth}
\begin{appendixprompt}{Coding-agent integration prompt}
\begin{AppendixPromptText}
Implement VisionSI in the supplied SI or RSI host using Sec. 3 and Algorithm 1. You have the host source code and the VQA training split. Preserve the host's proposal, mutation, archive, parent-selection, self-editing, and budget rules. Reserve annotated boxes and test labels for evaluation.

1. Output interface. Return ANSWER and EVIDENCE for each execution. EVIDENCE contains up to three [x1, y1, x2, y2] boxes with coordinates normalized to [0,1000]. Assign zero overlap when boxes are missing or unparsable.

2. Precomputation. Before host search, form groups from the training questions and cache each training question's answer-conditioned reference M_i and reliability weight w_i. Use the same references and weights for every candidate.

3. Candidate scoring. Identify where the host computes and uses candidate scores. Replace each per-example answer score with

    (A_i + w_i * IoU(B_i, M_i)) / 2

Use the host's existing aggregation and proposer feedback. In GEPA, update the evaluation adapter used for minibatch acceptance and validation Pareto selection. In Hyperagents, update staged and full evaluations and the scores passed to the archive and meta-agent. Keep credit computation outside the editable meta-agent artifact.

4. Final harness. Keep one harness from host search. Use the training groups from Step 2 to propose instructions for that harness and retain updates following Algorithm 1. Store the instructions in the same harness. At evaluation, select the instruction associated with the nearest training-group centroid.

5. Verification. Check that setting all w_i to zero recovers the answer-only ranking. Verify that optimization reads neither annotated boxes nor test data. In a smoke test, report A_i, IoU(B_i,M_i), w_i, and the combined score for three candidates on ten training questions.
\end{AppendixPromptText}
\end{appendixprompt}
\captionsetup{hypcap=false}
\captionof{figure}{\textbf{Coding-agent integration prompt for integrating \ours{} with existing SI and RSI algorithms.} The prompt follows Algorithm~\ref{alg:integration} and
identifies the scoring interfaces in GEPA and Hyperagents.}
\label{fig:integration_prompt}
\end{minipage}
\Needspace{.46\textheight}
\section{Additional Results}
\label{app:additional_results}

\subsection{Further Analysis of Evidence Drift}
\label{app:drift}

\begin{wrapfigure}{r}{.43\linewidth}
  \centering
  \includegraphics[width=\linewidth]{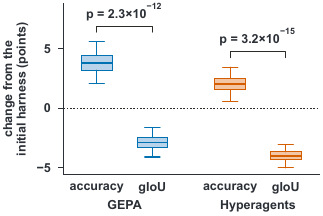}
  \caption{\textbf{Changes in accuracy and evidence localization after harness optimization.} Each change is measured against the initial harness and pooled over four VLMs on UMI-VQA and ScreenQA. Boxes span the middle 50\% of the bootstrap estimates of the mean change, and whiskers span 95\%. The bootstrap resamples questions within each VLM and domain. Brackets give a paired Wilcoxon test of whether, on the same questions, the accuracy change differs from the gIoU change.}
  \label{fig:drift_hosts}
\end{wrapfigure}
\paragraph{Accuracy can rise while localization worsens.}
We compare each host's optimized harness with the initial harness in all 20 combinations of four VLMs and five domains. After optimization, gIoU decreases in 11 of the 20 settings for each host. In 7 GEPA settings and 6 Hyperagents settings, this decrease occurs despite higher answer accuracy. Figure~\ref{fig:drift_hosts} shows UMI-VQA and ScreenQA, where Hyperagents lowers gIoU for all four VLMs. Across these two domains, GEPA gains 3.8 accuracy points and loses 2.9 gIoU points relative to the initial harness. Hyperagents gains 2.0 accuracy points and loses 4.0 gIoU points.

Most of GEPA's accuracy gain in this comparison comes from UMI-VQA, where its accuracy rises by 7.9 points. On ScreenQA, its accuracy changes by $-0.2$ points. EarthReason shows GEPA's largest localization loss among the five domains, with gIoU falling by 7.6 points while accuracy rises by 2.1 points.

\paragraph{Extending the blur diagnostic to RSI.}
The main paper uses whole-image blur to show that GEPA's accuracy gains can persist when visual information is degraded. We extend this diagnostic to Hyperagents to examine whether the same pattern appears under RSI. Table~\ref{tab:blur_hosts} reports results with Qwen3.5-2B across five domains. At 4\% blur, GEPA retains 6.1 points of its 7.2 point accuracy gain, while Hyperagents retains 3.6 points of its 6.1 point gain. Thus, both hosts retain part of their answer improvements under visual degradation, consistent with the limitation of answer-only optimization discussed in Sec.~\ref{sec:analysis}.

\begin{table}[htbp]
\caption{\textbf{Answer accuracy under whole-image blur.} Results use Qwen3.5-2B and the
blur radius is 4\% of the longer image side. Parentheses show the gain over the initial harness.}
\label{tab:blur_hosts}
\centering
\scriptsize
\setlength{\tabcolsep}{6pt}
\renewcommand{\arraystretch}{1.05}
\begin{tabular}{@{}lcc@{}}
\toprule
Harness & Original image & 4\% blur \\
\midrule
Initial harness & 41.7 & 19.7 \\
GEPA & 48.9 $(+7.2)$ & 25.9 $(+6.1)$ \\
Hyperagents & 47.9 $(+6.1)$ & 23.3 $(+3.6)$ \\
\bottomrule
\end{tabular}
\end{table}

\Needspace{.33\textheight}
\begin{wrapfigure}{r}{.39\linewidth}
  \centering
  \includegraphics[width=\linewidth]{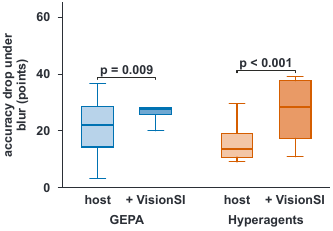}
  \caption{\textbf{Accuracy drop under whole-image blur on UMI-VQA with and without \ours{}.} The drop is test accuracy
  on the original image minus accuracy at 4\% blur. Each box summarizes the four VLMs. The line marks the median, the box
  spans the quartiles, and the whiskers mark the minimum and maximum. Brackets give two-sided paired bootstrap tests that
  resample images within each VLM.}
  \label{fig:blur_umi}
\end{wrapfigure}
\paragraph{Image dependence under visual degradation}
We compare the hosts and their \ours{} variants under whole-image blur on UMI-VQA across four VLMs. Figure~\ref{fig:blur_umi} shows that \ours{} has a 4.9 point larger accuracy drop than GEPA and a 10.2 point larger drop than Hyperagents when the images are blurred by 4\%. 

Greater image dependence does not require removing textual information from the harness. A textual instruction that tells the agent to locate a drawer can help it identify relevant visual evidence, while an instruction that simply supplies the answer may produce a correct prediction without locating the object. Both types of information can emerge during self-improvement, and the same textual cue may be useful for some questions but misleading for others. \ours{} therefore preserves textual guidance and focuses on improving the localization of reported evidence, without attempting to distinguish or eliminate every textual shortcut. The blur results suggest increased sensitivity to visual information, while the localization experiments directly measure the evidence drift that \ours{} is designed to mitigate.


\subsection{Analysis of Reliability-Aware Visual Credit}
\label{app:credit}

\paragraph{\ours{} improves localization on shared correct answers.}
We examine whether \ours{} reports more accurate evidence even when the host and its \ours{} variant both answer correctly. We evaluate gIoU on the shared set of correctly answered test questions, averaging equally across 15 combinations of three VLMs and five domains (randomly sampled for computational efficiency). On these questions, mean gIoU improves from 42.0 to 54.3 for GEPA and from 45.8 to 53.8 for Hyperagents when combined with \ours{}. These results show that the localization gains cannot be explained by differences in answer correctness on the evaluated questions.

\paragraph{Reliability weighting improves localization.}
We compare reliability-aware weighting with uniform weighting to evaluate the contribution of $w_i$. Both variants use the same model-generated references, but uniform weighting sets $w_i=1$ for every question. On UMI-VQA with Qwen3.5-2B, the training-set weights correlate with reference quality, measured by IoU against annotated regions with Spearman's $\rho=0.71$. The ground-truth bounding boxes are used only for this analysis, not for optimization. Across three optimization seeds per host, reliability weighting improves test gIoU by 2.8 points and cIoU by 3.6 points over uniform weighting. Figure~\ref{fig:w_ablation} shows that the gains are largest on low-weight questions. For these questions, the visual interventions indicate either weak image reliance or limited coverage of the relevant visual evidence by the reference. Uniform weighting gives full visual credit to agreement with these references, while reliability-aware weighting reduces their influence on harness optimization.

\Needspace{.40\textheight}
\subsection{Motivation for the Visual-Coherent Batch Sampler}
\label{app:sampling}

\begin{wrapfigure}{r}{.47\linewidth}
  \centering
  \includegraphics[width=\linewidth]{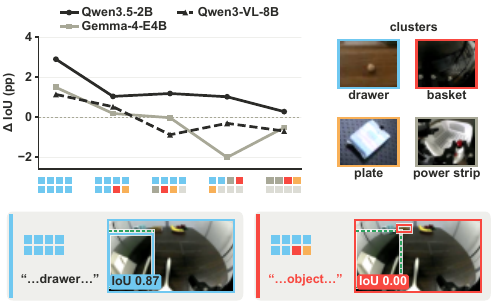}
  \caption{\textbf{Effect of batch composition on evidence localization.} We progressively introduce questions from different visual groups into an eight-question instruction-writing batch. Lines show the change in held-out gIoU for three VLMs on UMI-VQA, and colored squares indicate batch composition. The examples compare instructions and reported boxes generated from a visually coherent batch and a mixed batch.}
  \label{fig:scope_control}
\end{wrapfigure}
Visual instructions are more useful when they describe what to look for in a particular type of question. We investigate whether mixing questions with different visual targets makes these instructions less specific. Starting with eight questions from the same visual group, we progressively replace questions with examples from other groups. We generate an instruction from each batch and evaluate it on held-out questions from the original group, keeping the answer prompt fixed. Figure~\ref{fig:scope_control} shows that localization generally worsens as unrelated visual targets are introduced, with lower gIoU across all three tested VLMs on UMI-VQA. In the illustrated example, the single-group batch produces an instruction to look for the drawer, while the mixed batch produces a generic instruction referring to an object and reports an incorrect box. These results motivate the Visual-Coherent Batch Sampler, which groups questions with similar visual requirements to help preserve the specificity of learned instructions.

\subsection{Optimization Cost}
\label{app:cost}


\paragraph{Cost.}
\ours{} preserves the host's native proposal budget, so host search proposes the same number of candidate harnesses with and without \ours{}. Additional computation occurs both before and after host search. Before search, \ours{} constructs visual-coherent groups, generates answer-conditioned reference regions, and estimates their reliability using visual interventions. The references and weights are cached for candidate evaluation. After host search, an auxiliary instruction-learning stage generates and evaluates group-specific visual instructions for the selected harness. Averaged across model and domain settings, one \ours{} run uses approximately 65k VLM calls and 66M prompt tokens with GEPA, and 48k calls and 44M prompt tokens with Hyperagents. 

\subsection{Qualitative Results}
\label{app:qualitative}

\begin{figure}[!htbp]
\begin{minipage}[t]{.48\linewidth}
  \vspace{0pt}
  \captionsetup{hypcap=false}
  \centering
  \includegraphics[width=\linewidth]{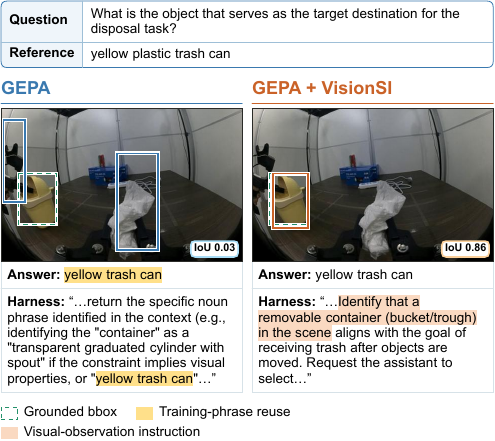}
  \captionof{figure}{\textbf{Evidence localization on UMI-VQA.} Both harnesses use Qwen3.5-2B with GEPA as the host, and both answer
  correctly. The GEPA harness holds the training phrase ``yellow trash can'', highlighted in yellow. Its answer repeats the
  phrase, and its boxes miss the annotated evidence. The harness optimized with \ours{} holds a visual-observation
  instruction, highlighted in orange, and boxes the trash can.}
  \label{fig:case_card_umi}
\end{minipage}\hfill
\begin{minipage}[t]{.48\linewidth}
  \vspace{0pt}
  \captionsetup{hypcap=false}
  \centering
  \includegraphics[width=\linewidth]{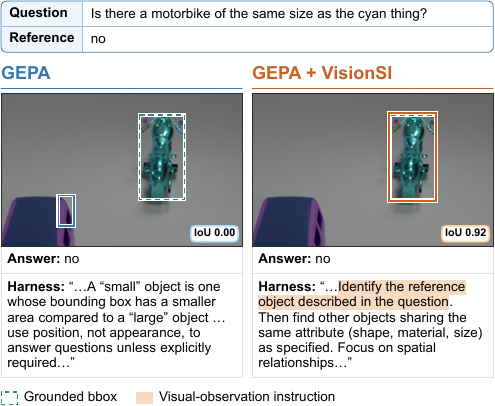}
  \captionof{figure}{\textbf{Evidence localization on Super-CLEVR.} Both harnesses use Qwen3-VL-8B with GEPA as the host, and both
  answer correctly. The GEPA harness tells the executor to judge size from box area and to use position rather than
  appearance. Its box falls on the van. The harness optimized with \ours{} instructs the executor to identify the reference
  object in the question, and its box covers the cyan motorbike.}
  \label{fig:case_card_superclevr}
\end{minipage}
\end{figure}

\begin{figure}[!htbp]
  \centering
  \includegraphics[width=.50\linewidth]{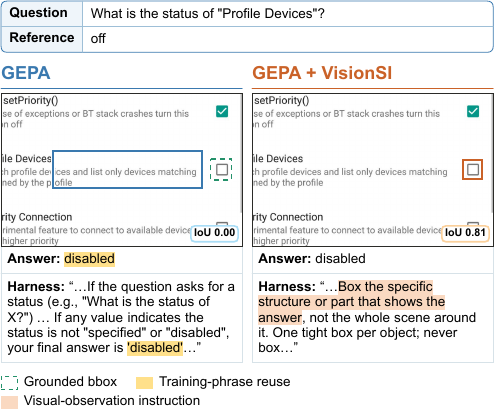}
  \caption{\textbf{Evidence localization on ScreenQA.} Both harnesses use Qwen3.5-2B with GEPA as the host, and both answer
  correctly. The GEPA harness maps status questions to the training answer ``disabled''. Its box covers the setting's
  description instead of its checkbox. The harness optimized with \ours{} instructs the executor to box the interface
  element that shows the answer. Its box covers the checkbox.}
  \label{fig:case_card_screenqa}
\end{figure}

\begin{figure}[!htbp]
  \centering
  \includegraphics[width=.73\linewidth]{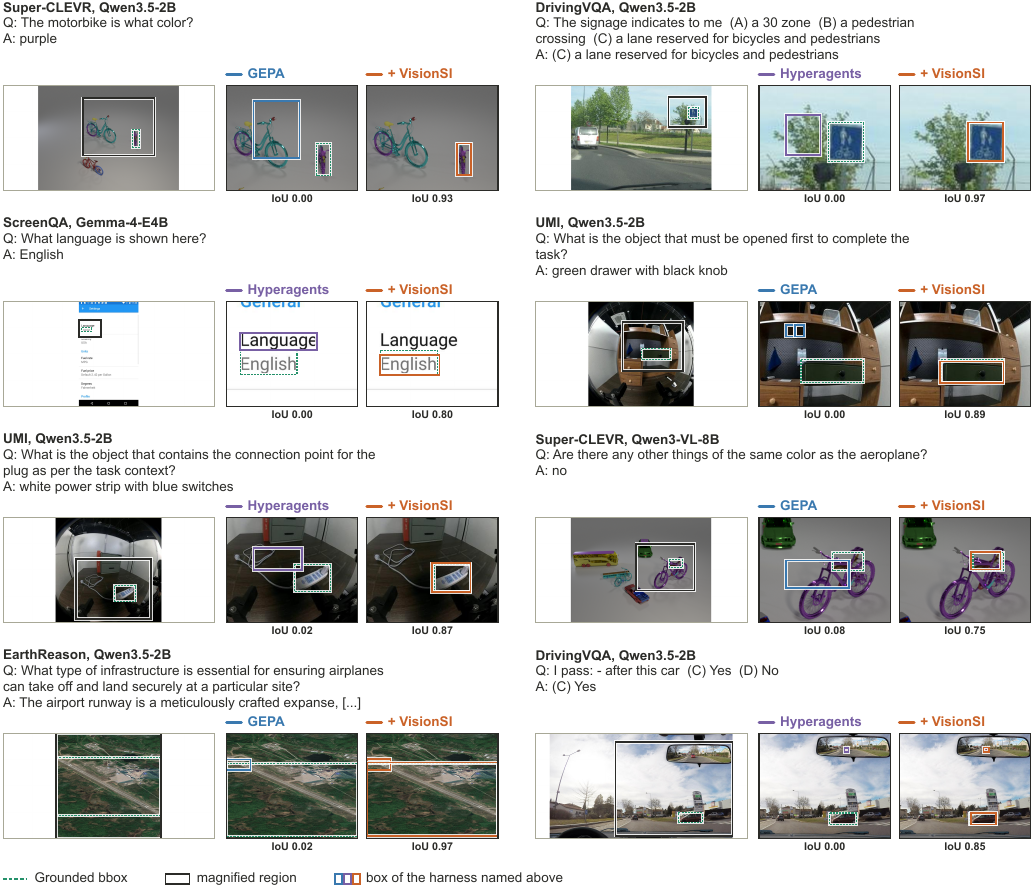}
  \caption{\textbf{Localization gains among correct answers.} Both harnesses answer correctly in each example. Dashed green boxes mark annotations, and black boxes
  mark the magnified regions.}
  \label{fig:case_appendix}
\end{figure}

Figures~\ref{fig:case_card_umi} to~\ref{fig:case_card_screenqa} show questions that both harnesses answer correctly but localize differently, together with the harness text behind each answer. On UMI-VQA and ScreenQA, the GEPA answer repeats a phrase stored in its harness.

On Super-CLEVR, both harnesses first box the objects and can refine the boxes on a magnified view, but their instructions point the executor to different objects. Across these three cases, the GEPA boxes have an IoU of at most 0.25, and the boxes from \ours{} have an IoU of at least 0.5.

Figure~\ref{fig:case_appendix} shows more examples with the same pattern. In each, both harnesses answer correctly, and the boxes from \ours{} overlap the annotated evidence more closely.

\section{Discussion}
\label{app:discussion}

Self-improvement can turn textual shortcuts into persistent instructions. Unlike the language prior of a frozen VLM, an optimized harness can retain shortcuts learned from previous executions and reuse them on future questions. Our blur analysis and case studies show that improvements in answer accuracy can coexist with reported boxes that miss the relevant visual evidence. There are two limitations of answer-only selection. It does not evaluate the reported visual evidence, and it depends on the reliability of the answer scorer. In RSI, these problems can also affect later generations because retained harnesses influence subsequent improvement steps.

Textual instructions can guide visual observation, but they can also introduce shortcuts. For example, an instruction to inspect a traffic sign's shape and symbols helps the agent identify relevant image evidence. In contrast, storing a frequent answer in the harness may improve answer accuracy while leaving the reported evidence misplaced. The same textual cue can be useful for some questions and misleading for others, making it difficult to distinguish task-relevant guidance from spurious shortcuts. \ours{} therefore retains textual guidance as part of harness optimization and uses reliability-aware visual credit to evaluate the evidence reported by candidate harnesses.

\ours{} thus addresses this problem by incorporating reliability-aware visual credit into harness selection. The localization gains on questions answered correctly by both methods in Sec.~\ref{app:credit} show that these improvements cannot be explained by answer accuracy alone. Visual interventions estimate how much credit a model-generated reference should receive, and the benefit over uniform weighting is largest on low-weight questions. However, a useful reference may receive a low weight when multiple regions contain similar evidence, since blurring one region may leave enough information to answer correctly. The Visual-Coherent Batch Sampler addresses a separate problem by grouping questions with similar visual requirements. Our mixed-batch experiment shows that adding unrelated questions can turn a specific instruction, such as looking for a drawer, into a generic instruction that provides less useful visual guidance.

Reported evidence gives an agent's answer a location that can be checked against the image. Even when the answer is correct, a box pointing to an unrelated object provides misleading evidence. During self-improvement, such errors can persist when the harness is reused on later questions. \ours{} addresses this problem by improving the localization of reported evidence during harness optimization. A separate question is whether the reported regions also contribute to the model's answers. This could be investigated in future studies.